\documentclass{article}

\PassOptionsToPackage{numbers,compress}{natbib}
\usepackage[preprint]{neurips_2024}

\usepackage{wrapfig}
\usepackage[utf8]{inputenc} % allow utf-8 input
\usepackage[T1]{fontenc}    % use 8-bit T1 fonts
\usepackage{hyperref}       % hyperlinks
\hypersetup{hidelinks}
\usepackage{url}            % simple URL typesetting
\usepackage{booktabs}       % professional-quality tables
\usepackage{pifont}         % for \ding symbols
\usepackage{amsfonts}       % blackboard math symbols
\usepackage{nicefrac}       % compact symbols for 1/2, etc.
\usepackage{microtype}      % microtypography
\usepackage[table]{xcolor}
\usepackage{lipsum}
\usepackage{graphicx}
\usepackage{times}
\usepackage{amsmath,amssymb}
\usepackage{enumitem}
\usepackage{xspace}
\usepackage{pifont} % for the \ding
\usepackage{multirow, makecell, tabularx}
\newcolumntype{C}{>{\centering\arraybackslash}X}
\newcolumntype{W}{>{\hsize=1.1\hsize\centering\arraybackslash}X} % wider

\usepackage{subcaption}
\usepackage{placeins}
\usepackage{algorithm}
\usepackage{algpseudocode}
\usepackage{tikz}
\usetikzlibrary{decorations.pathreplacing, shadows.blur, calc}

\newcommand{\method}{GEM-3\xspace}

\newcommand{\verify}[1]{\textcolor{red}{Verify.}}

\newcommand{\cmark}{\ding{51}}%
\definecolor{lightgray}{gray}{0.75}
\newcommand{\xmark}{\textcolor{lightgray}{\ding{55}}}% X mark symbol
\definecolor{pastelblue}{RGB}{70, 70, 70} % Light Blue

\usepackage{booktabs,tabularx}
\usepackage[most]{tcolorbox}
\definecolor{greydark}{RGB}{90, 90, 90} 
\definecolor{greylight}{RGB}{245, 245, 245}

\newtcolorbox{greycustomblock}{
  colframe=greydark,        % Left line color
  colback=greylight,        % Background color
  boxrule=1pt,              % Line width
  left=2.5pt,               % Inner left margin
  right=3pt,                % Inner right margin
  top=5pt,                  % Inner top margin
  bottom=3pt,               % Inner bottom margin
  arc=0pt,                  % No rounded corners
  breakable,                % Allow breaking across pages
  before skip=0.2\baselineskip, % Vertical space before
  after skip=0.2\baselineskip,  % Vertical space after
  left skip=0pt,            % Left skip (adjust as needed)
  right skip=0pt,           % Right skip (adjust as needed)
  enhanced jigsaw,          % Enhanced jigsaw for precise control
  frame hidden,             % Hide the default frame
  overlay={                 % Custom overlay for left line
    \draw[greydark, line width=2pt]
      ([yshift=-1pt]frame.north west) -- ([yshift=1pt]frame.south west); % Adjusting the y coordinates to match exactly
  },
  fontupper=\selectfont, % Font size and style for the table
}

\newtcolorbox{roundgrey}{
  colframe=greydark,        % Frame color
  colback=greylight,        % Background color
  boxrule=0.8pt,            % Frame line width
  left=5pt,                 % Inner left margin
  right=5pt,                % Inner right margin
  top=5pt,                  % Inner top margin
  bottom=5pt,               % Inner bottom margin
  arc=4pt,                  % Rounded corners
  breakable,                % Allow breaking across pages
  before skip=0.2\baselineskip,
  after skip=0.2\baselineskip,
  left skip=0pt,
  right skip=0pt,
  enhanced,                 % Enable advanced styling
  frame hidden=false,       % Show normal frame
  fontupper=\selectfont,    % Normal font inside
}

\usepackage[most]{tcolorbox}
\usepackage{titling} % lets us suppress the normal \maketitle if we want
\usepackage{setspace} % for slight looseness if you want

\definecolor{frontbg}{RGB}{247,249,252}      % big box fill
\definecolor{frontborder}{RGB}{210,215,225}  % big box border
\definecolor{fronttext}{RGB}{20,20,20}       % near-black text
\definecolor{frontsub}{RGB}{90,90,90}        % subtitle/affiliation grey

\newtcolorbox{frontpagebox}{
  breakable,
  colback=frontbg,
  colframe=frontborder,
  boxrule=0.5pt,
  arc=10pt,              % <-- rounded corners
  left=16pt,
  right=16pt,
  top=16pt,
  bottom=16pt,
  before skip=0pt,
  after skip=1.5em,
  enhanced,
}

\newtcolorbox{metadatabox}{
  colback=white,
  colframe=frontborder,
  boxrule=0.5pt,
  arc=8pt,
  left=10pt,
  right=10pt,
  top=8pt,
  bottom=8pt,
  enhanced,
}

\usepackage{fontawesome}
\usepackage{adjustbox} % Add this line to your preamble
\usepackage{tikz}

\newcommand{\lightbulbicon}{%
  \begin{tikzpicture}[baseline=-0.5ex]
    \draw[fill=white, draw=insightteal, thick] (0,0) circle (1.5ex);
    \node[scale=0.8, color=insightteal] at (0,0) {\faLightbulbO~};
  \end{tikzpicture}%
}

\definecolor{insightteal}{RGB}{34, 139, 139}   % A sophisticated, medium-dark teal
\definecolor{insightback}{RGB}{240, 248, 248}   % A very light, complementary cyan-white

\newtcolorbox{customblockquote}{
  colframe=insightteal,
  colback=insightback,
  boxrule=0pt,
  left=5pt,  % Set to 0pt so the background color touches the left 
  right=4pt,
  top=5pt,
  bottom=3pt,
  arc=0pt,
  breakable,
  before skip=1.2\baselineskip,
  after skip=0.7\baselineskip,
  left skip=0pt,
  right skip=0pt,
  enhanced jigsaw,
  frame hidden,
   overlay={
    \draw[insightteal, line width=2pt] 
      ([yshift=1pt]frame.north west) -- (frame.south west);
    \node[inner sep=0pt] at ([xshift=0pt, yshift=-1.3pt]frame.north west) {\lightbulbicon};
  },
  fontupper=\fontfamily{lmr}\selectfont,
  boxsep=1pt,
}

\newcommand{\fronttitle}[1]{%
  {\sffamily\bfseries\LARGE\raggedright #1\par}
}

\newcommand{\frontauthors}[1]{%
  {\sffamily\normalsize\textbf{#1}\par}
}

\newcommand{\frontabstracttext}[1]{%
  {\small #1\par}
}

\usepackage{titletoc}

\titlecontents{section}[0em]
{\addvspace{0.3em}\sffamily\bfseries\small\color{fronttext}}
{\thecontentslabel\hspace{1em}}
{\hspace*{-1em}}
{\titlerule*[0.7pc]{\textcolor{frontborder}{.}}\thecontentspage}

\titlecontents{subsection}[2.5em]
{\addvspace{0pt}\sffamily\footnotesize\color{greydark}}
{\thecontentslabel\hspace{1em}}
{\hspace*{-1em}}
{\titlerule*[0.7pc]{\textcolor{frontborder}{.}}\thecontentspage}

\newtcolorbox{executiveTOC}{
  colback=white,
  colframe=frontborder,
  boxrule=0.5pt,
  arc=6pt,
  left=20pt, right=20pt, top=14pt, bottom=14pt,
  before skip=1em,
  after skip=2em,
  enhanced,
}
\newenvironment{frontabstract}{\setlength{\parindent}{0pt}}{}

\begin{document}

\thispagestyle{empty}  % no header/footer number on first page

\begin{frontpagebox}
  \begin{frontabstract}

    % --- Title
    % \fronttitle{Forecasting Across Scales with One Model}
    %\fronttitle{GEM 3.0: Skillful and Efficient Multi-Scale Global Weather Forecasting}
    \fronttitle{Timestep-Conditioned Transformers for Global Weather Forecasting}

    \vspace{0.6em}

    % --- Authors / Team line
    \frontauthors{Salient}

    {\sffamily\footnotesize\color{frontsub}%
      Sam Levang, Fran Bartoli\'{c}, Ty Dickinson, Chase Dwelle, Paulius Rauba, Viktor Cikojevi\'{c}\\
    \par}

    \vspace{1.2em}

    % --- Execute-summary paragraph 
    \frontabstracttext{
    \vspace{-5mm}

Existing machine-learning weather forecasting models rely on predetermined and fixed autoregressive timesteps. The choice of model timestep involves a fundamental trade-off: shorter timesteps (e.g. 1 to 6 hours) finely resolve atmospheric dynamics within the diurnal cycle but increase error accumulation for a given forecast horizon, while longer timesteps (e.g. 24 hours) reduce error accumulation but limit the usability of short-range forecasts where sub-daily predictability is high. In this work, we present \method, a probabilistic global weather model that addresses this trade-off through explicit multi-timestep inference. With a single set of trained weights, the model timestep can be configured at inference time to balance predictability and usability across a broad forecast horizon. Additionally, we find that mixed-timestep training consistently improves rollout stability relative to timestep-specialist models. Under the hood, \method is a lightweight neighborhood-attention transformer with $\sim$134M parameters on an equirectangular grid with a number of architectural advancements beyond its predecessor GEM-2 \citep{rauba2026probabilistic}. The result is a practical forecasting system that couples near-SOTA medium-range probabilistic skill, stable extended-range rollouts, efficient training and inference, and decision-relevant diagnostics.
    
    \vspace{0.1em}

    }

    % --- metadata card, like Meta's "Date / Code / Weights ..."
    \begin{metadatabox}
      {\footnotesize
        \textbf{Date:} June 12, 2026\\
        \textbf{Contact:} \texttt{\{initial.surname\}@salientpredictions.com}
      }
    \end{metadatabox}

  \end{frontabstract}
\end{frontpagebox}

%\newpage

\section{Introduction}
\label{sec:intro}

\begin{figure*}[h]
\centering
% --- Panel (a): full-width timestep-schedule schematic ---
\resizebox{\textwidth}{!}{% Panel (a) of Figure 1: configurable inference-timestep schedule.
\begin{tikzpicture}[x=0.52cm, y=0.48cm]

% === Professional palette ===
\definecolor{cA}{RGB}{55,126,184}     % blue  - 6 h
\definecolor{cB}{RGB}{217,150,45}     % amber - 24 h
\definecolor{cG}{RGB}{30,145,65}      % green - hybrid / production
\definecolor{cX}{RGB}{80,80,90}       % neutral grey

\node[font=\small\sffamily\bfseries, text=black!78, anchor=south west]
    at (-2.6,5.6) {\textbf{a}};
\node[font=\tiny\sffamily, text=cX!70, anchor=south west]
    at (-1.4,5.65) {Fixed vs.\ configurable timestep schedules};

% --- Time axis ---
\draw[cX!50, line width=0.5pt, -stealth] (-0.2,-0.80) -- (21.2,-0.80);
\node[font=\tiny\sffamily, text=cX!75] at (0,-1.18) {$t_0$};
\node[font=\tiny\sffamily, text=cX!75] at (20,-1.18) {$\tau_{\max}$};
\node[font=\tiny\sffamily, text=cX!55] at (10,-1.18) {forecast horizon};

% === Row 1: Delta t = 6 h ===
\node[font=\scriptsize\sffamily, text=cA!90, anchor=east]
    at (-0.6,4.3) {$\Delta t{=}6\,\text{h}$};
\foreach \i in {0,...,19}{
    \fill[cA!32, rounded corners=0.5pt]
        (\i+0.07,3.85) rectangle (\i+0.93,4.75);
    \draw[cA!60, rounded corners=0.5pt, line width=0.3pt]
        (\i+0.07,3.85) rectangle (\i+0.93,4.75);
}
\node[font=\tiny\sffamily\itshape, text=cA!80, anchor=west, align=left]
    at (20.55,4.3) {more detail,\\deep rollout};

% === Row 2: Delta t = 24 h ===
\node[font=\scriptsize\sffamily, text=cB!90, anchor=east]
    at (-0.6,2.3) {$\Delta t{=}24\,\text{h}$};
\foreach \i in {0,...,4}{
    \fill[cB!32, rounded corners=0.5pt]
        (\i*4+0.07,1.85) rectangle (\i*4+3.93,2.75);
    \draw[cB!60, rounded corners=0.5pt, line width=0.3pt]
        (\i*4+0.07,1.85) rectangle (\i*4+3.93,2.75);
}
\node[font=\tiny\sffamily\itshape, text=cB!85, anchor=west, align=left]
    at (20.55,2.3) {less detail,\\shallow rollout};

% === Row 3: hybrid (production) ===
\fill[cG!12, rounded corners=2pt]
    (-0.15,-0.45) rectangle (20.25,0.85);
\draw[cG!50, rounded corners=2pt, line width=0.45pt, dashed]
    (-0.15,-0.45) rectangle (20.25,0.85);

\node[font=\scriptsize\sffamily\bfseries, text=cG!85, anchor=east]
    at (-0.6,0.20) {hybrid};

\node[font=\tiny\sffamily\itshape, text=cG!82, anchor=west, align=left]
    at (20.55,0.20) {lead-time\\optimized};

% 6 h segment (days 0-14): 7 small boxes
\foreach \i in {0,...,6}{
    \fill[cA!35, rounded corners=0.5pt]
        (\i+0.07,-0.15) rectangle (\i+0.93,0.55);
    \draw[cA!65, rounded corners=0.5pt, line width=0.35pt]
        (\i+0.07,-0.15) rectangle (\i+0.93,0.55);
}
% transition arrow
\draw[cG!75, line width=0.7pt, ->, >=stealth]
    (7.05,0.20) -- (7.88,0.20);
% 24 h segment (day 14 -> T): 3 wide boxes
\foreach \i in {0,1,2}{
    \fill[cB!32, rounded corners=0.5pt]
        (8+\i*4+0.07,-0.15) rectangle (8+\i*4+3.93,0.55);
    \draw[cB!60, rounded corners=0.5pt, line width=0.35pt]
        (8+\i*4+0.07,-0.15) rectangle (8+\i*4+3.93,0.55);
}

% Inline cadence tags for the hybrid row
\node[font=\tiny\sffamily, text=cA!85, anchor=center]
    at (3.5,1.25) {6\,h};
\node[font=\tiny\sffamily, text=cB!88, anchor=center]
    at (14,1.25) {24\,h};

\end{tikzpicture}}

\vspace{1.0em}

% --- Panels (b) and (c): conditioning mechanism + timestep skill tradeoff ---
\begin{minipage}[t]{0.36\textwidth}
\raggedright
{\small\sffamily\bfseries\color{black!78}b}\,\,{\tiny\sffamily\color{black!40}Sequential transformer conditioning}\\[2pt]
\resizebox{\linewidth}{!}{% Panel (b) of Figure 1: simplified multi-timestep conditioning mechanism.
% A condensed teaser of the full architecture in Fig.~\ref{fig:architecture}.
\begin{tikzpicture}[x=1cm, y=1cm,
    panel/.style={fill=cM!2, draw=cM!28, rounded corners=4pt, line width=0.5pt},
    bandT/.style= {rounded corners=2pt, draw=cDT!75, fill=cDT!28, line width=0.7pt},
    bandZ/.style= {rounded corners=2pt, draw=cZ!75,  fill=cZ!26,  line width=0.7pt},
    plain/.style= {rounded corners=2pt, draw=cX!55,  fill=cX!12,  line width=0.7pt},
    src/.style  = {rounded corners=2pt, draw=#1!75,  fill=#1!22,  line width=0.7pt,
                   font=\scriptsize\sffamily, inner sep=2.2pt},
    io/.style   = {font=\scriptsize\sffamily, text=cX!90},
    comp/.style = {font=\scriptsize\sffamily, text=cX!95},
    fl/.style   = {-stealth, cX!65, line width=0.8pt},
    cfl/.style  = {-stealth, #1!85, line width=0.9pt},
]

\definecolor{cDT}{RGB}{217,150,45}   % amber - timestep
\definecolor{cZ}{RGB}{30,145,65}     % green - noise
\definecolor{cM}{RGB}{70,130,180}    % panel fill (matches Fig.~\ref{fig:architecture})
\definecolor{cX}{RGB}{80,80,90}      % neutral grey

\def\cx{1.95}

\node[io] (xin) at (\cx,6.65) {$X_t$};

% block container
\draw[panel] (0.55,2.65) rectangle (3.35,6.10);
\node[font=\tiny\sffamily, text=cM!70, anchor=north west] at (0.66,6.02) {transformer block};
\node[font=\tiny\sffamily\bfseries, text=cM!55, anchor=north east] at (3.24,6.02) {$\times L$};

% internal stack (top to bottom)
\node[plain, minimum width=2.15cm, minimum height=0.32cm, comp] (ln)  at (\cx,5.30) {LayerNorm};
\node[bandT, minimum width=2.15cm, minimum height=0.32cm, comp] (mt)  at (\cx,4.62)
      {$(1{+}\gamma_{\Delta t})\,\cdot\,{+}\,\beta_{\Delta t}$};
\node[bandZ, minimum width=2.15cm, minimum height=0.32cm, comp] (mz)  at (\cx,3.94)
      {$(1{+}\gamma_{z})\,\cdot\,{+}\,\beta_{z}$};
\node[plain, minimum width=2.15cm, minimum height=0.32cm, comp] (sub) at (\cx,3.20) {Attention\,/\,MLP};

\node[io] (xout) at (\cx,2.00) {$X_{t+\Delta t}$};

% flow arrows
\draw[fl] (xin) -- (\cx,6.10);
\draw[fl] (ln) -- (mt);
\draw[fl] (mt) -- (mz);
\draw[fl] (mz) -- (sub);
\draw[fl] (\cx,2.65) -- (xout);

% conditioning sources (left)
\node[src=cDT] (dt)  at (-0.85,5.30) {$\Delta t$};
\node[src=cDT] (fou) at (-0.85,4.62) {Fourier};
\draw[cfl=cDT] (dt) -- (fou);
\draw[cfl=cDT] (fou.east) -| (mt.west);

\node[src=cZ] (z) at (-0.85,3.94) {$z\!\sim\!\mathcal{N}$};
\draw[cfl=cZ] (z.east) -- (mz.west);

% order tags (right)
\node[font=\tiny\sffamily\bfseries, text=cDT!90, anchor=west] at (3.48,4.62) {1\textsuperscript{st}};
\node[font=\tiny\sffamily\bfseries, text=cZ!90,  anchor=west] at (3.48,3.94) {2\textsuperscript{nd}};

% compact legend (below), one signal per line
\node[bandT, minimum width=0.30cm, minimum height=0.18cm] (lT) at (0.05,1.35) {};
\node[font=\tiny\sffamily, text=cX!85, anchor=west] at (0.30,1.35)
      {$\Delta t$\,: timestep};
\node[bandZ, minimum width=0.30cm, minimum height=0.18cm] (lZ) at (0.05,0.88) {};
\node[font=\tiny\sffamily, text=cX!85, anchor=west] at (0.30,0.88)
      {$z$\,: noise vector};

\end{tikzpicture}}
\end{minipage}\hfill
\begin{minipage}[t]{0.58\textwidth}
\raggedright
{\small\sffamily\bfseries\color{black!78}c}\,\,{\tiny\sffamily\color{black!40}Timestep skill tradeoff (z500)}\\[2pt]
\includegraphics[width=\linewidth]{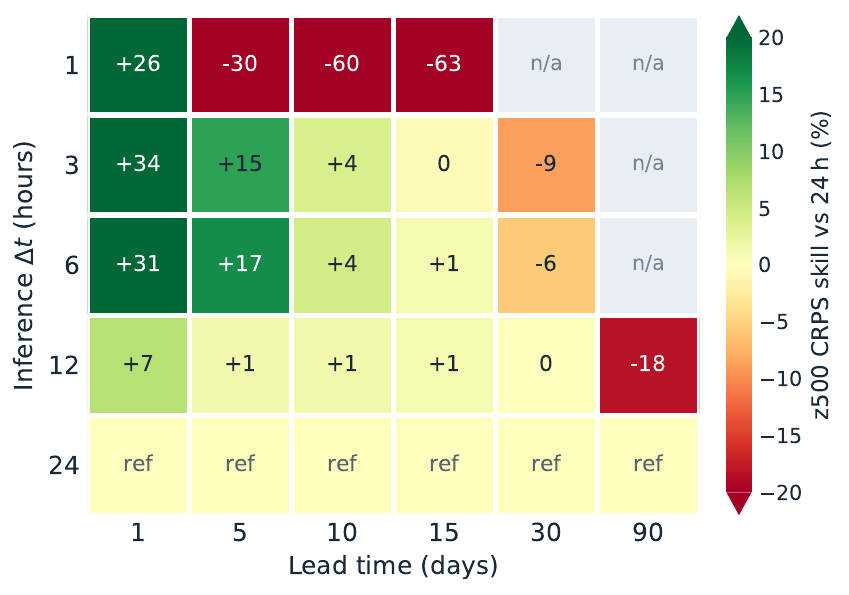}
\end{minipage}

\caption{\textbf{Configurable timestep inference with a single model.}
\textbf{(a)}~Finer timesteps of \textcolor[RGB]{55,126,184}{6\,h} or less resolve diurnal structure but require a deeper rollout to reach a given lead, while a coarse \textcolor[RGB]{217,150,45}{24\,h} schedule reaches it in fewer steps. The operational \method model uses a \textcolor[RGB]{30,145,65}{hybrid} schedule, running at 6\,h through day~14 and switching to 24\,h for the extended range.
\textbf{(b)}~A single set of weights supports this via AdaLN conditioning of the transformer blocks with a Fourier-embedded timestep $\Delta t$ in addition to an ensemble noise vector $z$. Full details in Fig.~\ref{fig:architecture}.
\textbf{(c)}~The resulting forecast skill tradeoff, shown as 500\,hPa geopotential CRPS skill relative to the 24\,h schedule for a single configurable 1--24\,h timestep generalist model (positive is better). Fine steps perform slightly better at short leads but degrade more quickly under rollout. Coarser steps remain stable to much longer leads, motivating the hybrid schedule.}
\label{fig:multitemporal}
\end{figure*}
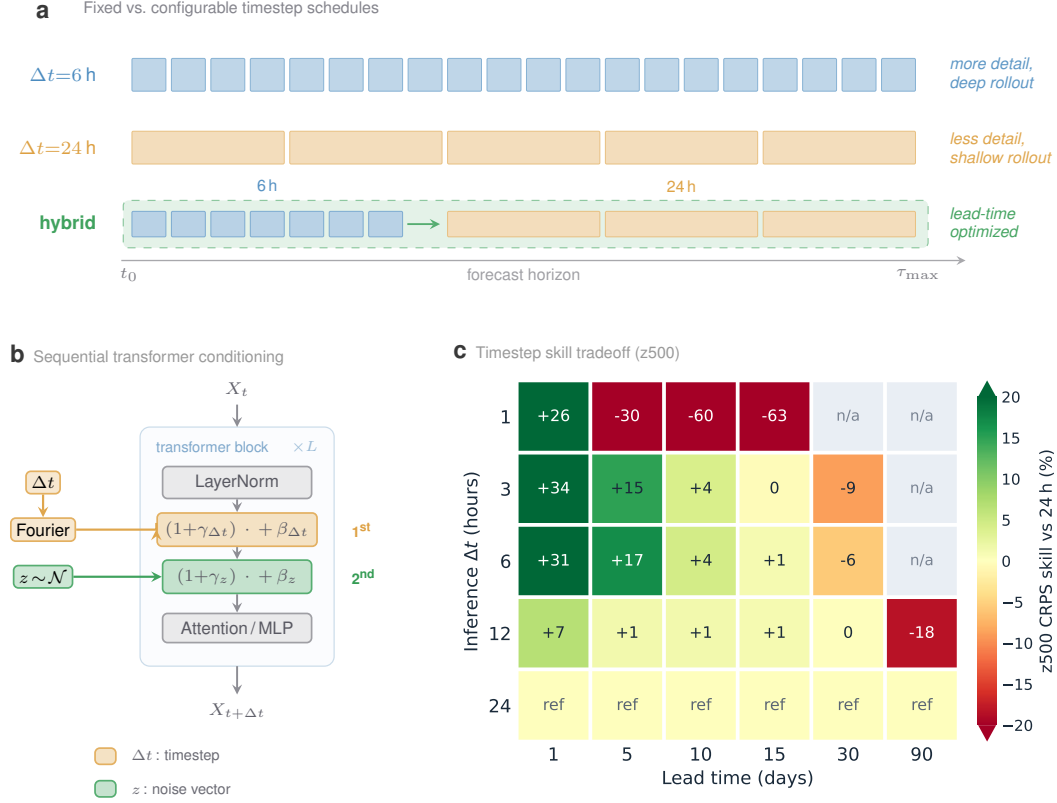

Machine-learning (ML) weather forecasting models now routinely exceed the skill of operational numerical weather prediction (NWP) systems across deterministic \citep{pathak2022fourcastnet, bi2023pangu, lam2023learning, chen2023fuxi, chen2025fengwu, keisler2022gnn}, probabilistic \citep{price2023gencast, price2024probabilistic, alet2025skillful, bonev2025fourcastnet3, lang2024aifs, couairon2024archesweather, weyn2021subseasonal}, and foundation-level \citep{nguyen2023climax, kochkov2024neuralgcm, sun2025fuxiweather, alexe2024graphdop} benchmarks. However, the architectural design of these atmospheric emulators imposes a limitation. High-frequency models (e.g., 1- to 6-hourly steps) finely resolve the temporal evolution of weather, but suffer from increased error accumulation in the autoregressive setting, and increased computational costs over extended horizons \citep{stock2025swift, nguyen2025omnicast}. Conversely, models operating natively at coarser temporal resolutions (e.g., 24-hourly) do not output the details of sub-daily trajectories and likely suffer from worse predictive performance for very coarse steps. As a result, practitioners must choose among models that differ in their capabilities: (i) models optimized for short lead times; (ii) models optimized for longer lead times that fail to capture short-term variability; or (iii) approaches that rely on post-processing to transform short-lead predictions into the required variables, which can introduce structural errors \citep{rauba2026probabilistic}.

\textbf{Multi-timestep inference}. In this work, we introduce an architectural mechanism that addresses this trade-off with configurable \textit{multi-timestep inference}. Existing global weather forecasting models perform training on either a single timestep size \citep{lam2023learning, price2023gencast, lang2024aifs, chen2023fuxi, bonev2025fourcastnet3}, or over a curriculum of timesteps but with no inference-time conditioning mechanism \citep{alet2025skillful}. In contrast, \method explicitly parameterizes the forecasting timestep $\Delta t$ as a conditioning variable. The timestep conditioning signal follows the same low-dimensional modulation pathway as the noise vector used to generate a stochastic ensemble. This formulation allows a single set of network weights to dynamically switch temporal resolutions at inference time.

\textbf{Anomaly-space modeling}. Because the intended use cases of \method include extended-range forecasting, maximizing autoregressive stability across the full rollout horizon is a high priority. Learned weather models do not close mass, heat, and moisture budgets exactly, and small imbalances can accumulate into drift over long horizons. To this end, we introduce a second architectural mechanism: anomaly-space modeling, in which we transform selected variables into climatological anomalies relative to an externally computed climatology. The core model then operates in a stable zero-centered residual state space with a more naturally attractive manifold. Rollout training and inclusion of temporal context features are also used to further improve stability.

\textbf{Architectural details}. \method operates on an equirectangular grid. Instead of the shifted-window transformer blocks of \citet{rauba2026probabilistic}, \method uses neighborhood attention (NATTEN) blocks \citep{hassani2023neighborhood}, which provide a more continuous locality bias. We couple this architecture with other recent improvements from modern transformer designs, including SwiGLU MLPs, QK-normalization, and a hybrid Muon \citep{liu2025muon, jordan2024muon} and AdamW optimization strategy. As with predecessor GEM models, we jointly model prognostic and diagnostic variables \citep{rauba2026probabilistic}.

\textbf{Results}. We demonstrate that \method achieves near-state-of-the-art continuous ranked probability scores (CRPS), outperforming ECMWF's ENS across all variables and lead times analyzed here. \method is also broadly comparable to other top-performing ML models such as AIFS-ENS \citep{lang2024aifs} and FGN \citep{alet2025skillful} on core dynamical benchmarks like z500, and has significant advantages on its native sub-timestep diagnostics like minimum and maximum temperature. \method also has stable and skillful extended-range rollouts, generally outperforming the ECMWF extended-range ensemble out to 46 days, and has good convergence towards climatological baseline scores out to 126 days.

\newpage
\begin{customblockquote}
\paragraph{Contributions.} In this report, we introduce a mechanism for \textit{multi-timestep inference} (Sec.~\ref{subsec:multi_timestep_inference}) which allows us to adaptively change the timestep of the autoregressive forecast within a single model at inference time. We use this mechanism plus a number of other advancements to build a skillful and stable multi-timestep inference model \method, with a public reforecast dataset available (Sec.~\ref{sec:global_model}).
\end{customblockquote}

\section{Problem setup}
\label{sec:setup}

\textbf{Preliminaries}. We work on a global latitude--longitude grid (equirectangular projection) with height $H_{\text{grid}}$ and width $W_{\text{grid}}$. Let \(X_t \in \mathbb{R}^{C_x \times H_{\text{grid}} \times W_{\text{grid}}}\) denote the atmospheric state at forecast lead $t$, represented as a multi-channel tensor of gridded variables (e.g., temperature, winds, pressure, humidity). Let \(C \in \mathbb{R}^{C_c \times H_{\text{grid}} \times W_{\text{grid}}} \) denote conditioning fields that constrain the evolution and are treated as given over the forecast window. Finally, let $X_{\le 0}$ denote the assimilated information available at initialization, which may include a window of past analyses and/or observations encoded into the initial model input. We write
\(
H := (X_{\le 0}, C) \in \mathcal{H}
\)
for the full information set available to the forecaster at initialization, with $\mathcal{H}$ the corresponding space. A particular forecast instance corresponds to a realized $H=h$.

\textbf{Modeling weather dynamics}. We assume the atmosphere induces an unknown conditional distribution over future trajectories given the initialization information. For a forecast horizon of $T$ leads, define the (unknown) conditional law
\begin{equation}
\label{eq:conditional_trajectory_law}
p(\,\cdot \mid h) \;:=\; \mathbb{P}\big(X_{1:T} \in \cdot \mid H=h\big),
\qquad
X_{1:T}\mid H=h \sim p(\,\cdot \mid h),
\end{equation}
where $X_{1:T} := (X_1,\dots,X_T)$. We also denote the lead-time marginals by
\begin{equation}
\label{eq:lead_time_marginal}
p_t(\,\cdot \mid h) \;:=\; \mathbb{P}\big(X_t \in \cdot \mid H=h\big).
\end{equation}
The goal of probabilistic forecasting is to represent these conditional distributions as accurately as possible.

\textbf{Approximating the model}. A forecasting model with parameters $\theta$ defines an approximate conditional distribution
\begin{equation}
\label{eq:model_trajectory_law}
q_\theta(\,\cdot \mid h) \;\approx\; p(\,\cdot \mid h),
\qquad
X_{1:T}\mid H=h \sim q_\theta(\,\cdot \mid h).
\end{equation}
Equivalently, the model specifies a family of conditional distributions over $X_t$ given the information available at prediction time (either the initialization information alone in a single-shot model, or the previously generated states in an autoregressive model). The central approximation target in this work is the full trajectory law $p(X_{1:T}\mid H)$, represented by $q_\theta(X_{1:T}\mid H)$.

\textbf{First-order Markov approximation}. We formulate temporal structure using a first-order Markov approximation in discrete time. With $X_0$ denoting the model input at initialization, including any encoded past context from $X_{\le 0}$, the approximation takes the form
\begin{equation}
\label{eq:first_order_markov}
p(X_t \mid X_{<t}, C) \;=\; p(X_t \mid X_{t-1}, C),
\qquad t=1,\dots,T.
\end{equation}
Under this approximation, the trajectory distribution factorizes as
\begin{equation}
\label{eq:true_markov_factorization}
p(X_{1:T} \mid H)
\;=\;
\prod_{t=1}^T p\bigl(X_t \mid X_{t-1}, C\bigr),
\end{equation}
where the dependence on $H$ enters through the initial model input $X_0$ and $C$. We parameterize the model $q_\theta$ to respect the same factorization.

\section{Multi-Timestep Inference}
\label{subsec:multi_timestep_inference}

Throughout this section, $t$ denotes physical forecast lead time, while $\Delta t$ denotes the time interval advanced by one model step. When we need to count autoregressive applications explicitly, we use $n$; after $n$ applications of a $\Delta t$-conditioned transition, the physical lead is $t_n=n\Delta t$. We use $\tau$ for a fixed physical lead when comparing different choices of $\Delta t$.
\subsection{Timestep tradeoff}
\label{subsec:tradeoff}

Most existing forecasting architectures commit to a fixed timestep $\Delta t$ at design and training time. This choice simultaneously fixes (i) the conditional transition law the model must approximate per step, and (ii) the rollout depth required to reach a given physical lead time. For leads $\tau$ that are integer multiples of $\Delta t$, we use $q_{\theta,\tau}^{\Delta t}(\cdot\mid H)$ to denote the marginal predictive distribution at physical lead $\tau$, induced by composing the $\Delta t$-conditioned transition $N(\tau,\Delta t)=\tau/\Delta t$ times. Under a first-order Markov rollout over a horizon $\tau=N\Delta t$, deploying at timestep $\Delta t$ induces a trajectory law of the form
\begin{equation}
\label{eq:fixed_timestep_trajectory}
q_\theta(X_{\Delta t},\ldots,X_{N\Delta t}\mid H,\Delta t)
\;=\;
\prod_{n=1}^{N} q_\theta\!\bigl(X_{n\Delta t} \mid X_{(n-1)\Delta t}, C, \Delta t\bigr),
\end{equation}
so the distribution at physical lead $\tau$ is obtained by repeated composition of the $\Delta t$-step transition.

For mixed schedules, the same construction applies with the product taken over the chosen sequence of step sizes. This coupling turns timestep selection into an objective trade-off that appears directly in expected forecast skill. For any proper scoring rule $S$ (e.g., CRPS), the expected score at physical lead $\tau$ under timestep $\Delta t$ is
\begin{equation}
\label{eq:expected_score}
\mathbb{E}_{H,\,X_\tau \sim p_\tau(\cdot\mid H)}
\Big[\, S\!\big(q_{\theta,\tau}^{\Delta t}(\cdot\mid H),\, X_\tau\big)\,\Big].
\end{equation}
Here the expectation is over forecast initializations $H$ and the corresponding verifying states $X_\tau$ drawn from $p_\tau(\cdot\mid H)$.
Short-range decision quality emphasizes small physical leads, where smaller $\Delta t$ typically improves this expectation because the $\Delta t$-step transition is closer to a local evolution operator and therefore easier to learn with high temporal fidelity. Long-range skill emphasizes larger physical leads, where the same deployment choice forces repeated application of an approximate kernel. In particular, the induced marginal at physical lead $\tau=N\Delta t$ can be written explicitly as
\begin{equation}
\label{eq:fixed_timestep_marginal}
q_{\theta,\tau}^{\Delta t}(X_\tau\mid H)
=
\int \prod_{n=1}^{N} q_\theta\!\bigl(X_{n\Delta t} \mid X_{(n-1)\Delta t}, C, \Delta t\bigr)\,\mathrm{d}X_{\Delta t}\cdots \mathrm{d}X_{(N-1)\Delta t},
\end{equation}
so any systematic mismatch in the per-step transition affects the final distribution through every factor in the product. Holding the physical horizon $\tau$ fixed, decreasing $\Delta t$ increases the number of factors $N(\tau,\Delta t)$ proportionally, which increases the number of times transition error is injected into the rollout and therefore degrades the long-lead expectation above through compounding. Increasing $\Delta t$ reduces the rollout depth and mitigates this accumulation mechanism, while simultaneously pushing the model toward a coarser, temporally aggregated transition that sacrifices short-range granularity.

As a consequence, \textit{short-lead and long-lead evaluations generally induce different preferred timesteps}: optimizing the expected score at small leads favors finer $\Delta t$, while optimizing the expected score at long leads favors coarser $\Delta t$ due to reduced compositional depth. Multi-timestep inference targets this incompatibility by conditioning a single parameter set $\theta$ on $\Delta t$, aiming to preserve short-range fidelity while maintaining long-range stability under rollout. 

\subsection{Joint Temporal and Aleatoric Modulation}
\label{subsec:joint_modulation}

We parameterize the one-step transition with an explicit timestep input and an explicit stochastic modulation input. Let $z \sim \mathcal{N}(0,I)$ be sampled independently for each ensemble member. The network maps
\begin{equation}
\label{eq:stochastic_transition_map}
\hat X_{t+\Delta t}
\;=\;
G_\theta(X_t, C, \Delta t, z),
\end{equation}
which induces the Markov kernel
\begin{equation}
\label{eq:conditioned_markov_kernel}
q_\theta(\,\cdot \mid X_t, C, \Delta t)
\;=\;
\operatorname{Law}\!\left(G_\theta(X_t, C, \Delta t, Z)\right),
\qquad Z\sim\mathcal{N}(0,I).
\end{equation}
The timestep $\Delta t$ specifies the physical transition scale the kernel is meant to emulate, while $z$ selects a stochastic realization from that conditional law, enabling ensemble diversity.

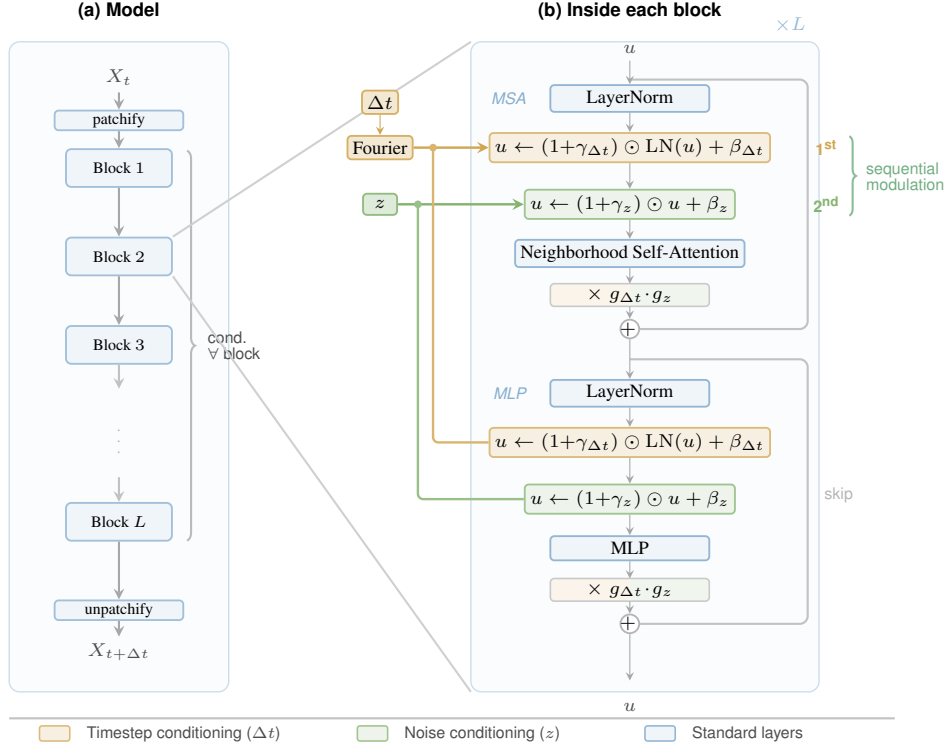
\begin{figure*}[t]
\centering
\begin{tikzpicture}[
    comp/.style={rectangle, draw=cM!58, fill=cM!8,
                 minimum width=2.1cm, minimum height=0.32cm,
                 font=\scriptsize, inner sep=2pt, rounded corners=1.5pt, line width=0.7pt},
    modT/.style={rectangle, draw=cT!65, fill=cT!14,
                 minimum width=2.1cm, minimum height=0.32cm,
                 font=\scriptsize, inner sep=2pt, rounded corners=1.5pt, line width=0.7pt},
    modN/.style={rectangle, draw=cN!65, fill=cN!12,
                 minimum width=2.1cm, minimum height=0.32cm,
                 font=\scriptsize, inner sep=2pt, rounded corners=1.5pt, line width=0.7pt},
    plus/.style={circle, draw=cX!55, fill=white,
                 minimum size=0.24cm, font=\tiny\bfseries, inner sep=0pt, line width=0.7pt},
    src/.style={rectangle, draw=#1!75, fill=#1!22,
                minimum height=0.28cm, font=\scriptsize,
                inner sep=2pt, rounded corners=1.5pt, line width=0.7pt},
    ob/.style={rectangle, draw=cM!50, fill=cM!8,
               minimum width=1.4cm, minimum height=0.5cm,
               font=\tiny, inner sep=2pt, rounded corners=2pt, line width=0.7pt},
    iob/.style={rectangle, draw=cM!50, fill=cM!8,
                minimum width=1.7cm, minimum height=0.26cm,
                font=\tiny, inner sep=1pt, rounded corners=1.5pt, line width=0.7pt},
    fl/.style={-stealth, cX!65, line width=0.8pt},
    cfl/.style={-stealth, #1!85, line width=0.9pt},
    sk/.style={cX!45, line width=0.8pt, rounded corners=3pt},
    lbl/.style={font=\scriptsize\sffamily\bfseries},
    ann/.style={font=\tiny\sffamily, text=black!70},
    panel/.style={fill=cM!2, draw=cM!28, rounded corners=4pt, line width=0.5pt},
]

\definecolor{cM}{RGB}{70,130,180}
\definecolor{cT}{RGB}{200,150,50}
\definecolor{cN}{RGB}{110,165,85}
\definecolor{cG}{RGB}{76,153,96}
\definecolor{cX}{RGB}{120,120,120}

% ===== (a) MODEL OVERVIEW =====
\def\modelcx{1.05}  % horizontal center of panel (-0.4 .. 2.5)
\draw[panel] (-0.4,0.0) rectangle (2.5,8.6);
\node[lbl] at (\modelcx,9.0) {(a) Model};

\begin{scope}[shift={(0,-0.12)}]
\node[ann, font=\scriptsize\sffamily] at (\modelcx,8.25) {$X_t$};
\node[iob] (patch) at (\modelcx,7.68) {patchify};
\node[ob] (b1) at (\modelcx,7.05) {Block $1$};
\node[ob] (b2) at (\modelcx,5.88) {Block $2$};
\node[ob] (b3) at (\modelcx,4.71) {Block $3$};
\node[font=\tiny, text=cX!55] at (\modelcx,3.54) {$\vdots$};
\node[ob] (bL) at (\modelcx,2.37) {Block $L$};
\node[iob] (unpatch) at (\modelcx,1.20) {unpatchify};
\node[ann, font=\scriptsize\sffamily] at (\modelcx,0.65) {$X_{t+\Delta t}$};

\draw[fl] (\modelcx,8.05) -- (patch);
\draw[fl] (patch) -- (b1);
\draw[fl] (b1) -- (b2);
\draw[fl] (b2) -- (b3);
\draw[fl,cX!45] (b3) -- (\modelcx,4.13);
\draw[fl,cX!45] (\modelcx,2.96) -- (bL);
\draw[fl] (bL) -- (unpatch);
\draw[fl] (unpatch) -- (\modelcx,0.85);

\draw[decorate, decoration={brace, amplitude=3pt}, cX!50, line width=0.8pt]
    (1.95,7.28) -- (1.95,2.12);
\node[ann, anchor=west, align=left] at (2.10,4.70) {cond.\\[-1pt]$\forall$ block};
\end{scope}

% ===== ZOOM INDICATOR =====
\draw[cX!35, line width=0.8pt] (b2.north east) -- (5.7,8.6);
\draw[cX!35, line width=0.8pt] (b2.south east) -- (5.7,0.0);

% ===== (b) TRANSFORMER BLOCK DETAIL =====
\node[lbl] at (7.8,9.0) {(b) Inside each block};

\draw[panel] (5.7,0.0) rectangle (10.3,8.6);
\node[font=\scriptsize\sffamily, text=cM!55, anchor=south east]
    at (10.25,8.6) {$\times L$};

\node[ann, font=\scriptsize\sffamily] at (7.8,8.5) {$u$};
\node[ann, font=\scriptsize\sffamily] at (7.8,-0.2) {$u$};

% --- MSA branch ---
% Skip (raw coordinates, drawn first so main flow is on top)
\draw[sk] (7.8,8.1) -| (10.15,4.8) -- (7.9,4.8);

% LN1 + input arrow that reaches LN1
\node[comp] (ln1) at (7.8,7.85) {LayerNorm};
\draw[fl, thin] (7.8,8.35) -- (ln1);

% Timestep modulation
\node[modT] (mt1) at (7.8,7.2)
    {$u \leftarrow (1{+}\gamma_{\Delta t})\odot \mathrm{LN}(u) + \beta_{\Delta t}$};
\draw[fl, thin] (ln1) -- (mt1);

% Noise modulation
\node[modN] (mn1) at (7.8,6.45)
    {$u \leftarrow (1{+}\gamma_{z})\odot u + \beta_{z}$};
\draw[fl, thin] (mt1) -- (mn1);

% Self-Attention
\node[comp, minimum width=2.75cm] (sa) at (7.8,5.8) {Neighborhood Self-Attention};
\draw[fl, thin] (mn1) -- (sa);

% Gate
\fill[cT!10, rounded corners=1.5pt] (6.75,5.1) rectangle (7.8,5.4);
\fill[cN!10, rounded corners=1.5pt] (7.8,5.1) rectangle (8.85,5.4);
\draw[cX!40, rounded corners=1.5pt, line width=0.6pt] (6.75,5.1) rectangle (8.85,5.4);
\node[font=\scriptsize] at (7.8,5.25)
    {$\times\; g_{\Delta t} \!\cdot\! g_{z}$};
\draw[fl, thin] (sa) -- (7.8,5.4);

% Plus (residual add)
\node[plus] (p1) at (7.8,4.8) {\scriptsize$\!+$};
\draw[fl, thin] (7.8,5.1) -- (p1);

% --- MLP branch ---
% Skip (raw coordinates, drawn first)
\draw[sk] (7.8,4.4) -| (10.15,0.9) -- (7.9,0.9);

% LN2 + arrow from p1 that reaches LN2
\node[comp] (ln2) at (7.8,3.95) {LayerNorm};
\draw[fl, thin] (p1) -- (ln2);

% Timestep modulation
\node[modT] (mt2) at (7.8,3.3)
    {$u \leftarrow (1{+}\gamma_{\Delta t})\odot \mathrm{LN}(u) + \beta_{\Delta t}$};
\draw[fl, thin] (ln2) -- (mt2);

% Noise modulation
\node[modN] (mn2) at (7.8,2.55)
    {$u \leftarrow (1{+}\gamma_{z})\odot u + \beta_{z}$};
\draw[fl, thin] (mt2) -- (mn2);

% MLP
\node[comp] (mlp) at (7.8,1.9) {MLP};
\draw[fl, thin] (mn2) -- (mlp);

% Gate
\fill[cT!10, rounded corners=1.5pt] (6.75,1.2) rectangle (7.8,1.5);
\fill[cN!10, rounded corners=1.5pt] (7.8,1.2) rectangle (8.85,1.5);
\draw[cX!40, rounded corners=1.5pt, line width=0.6pt] (6.75,1.2) rectangle (8.85,1.5);
\node[font=\scriptsize] at (7.8,1.35)
    {$\times\; g_{\Delta t} \!\cdot\! g_{z}$};
\draw[fl, thin] (mlp) -- (7.8,1.5);

% Plus (residual add)
\node[plus] (p2) at (7.8,0.9) {\scriptsize$\!+$};
\draw[fl, thin] (7.8,1.2) -- (p2);
\draw[fl, thin] (p2) -- (7.8,0.15);

% --- Conditioning sources (between panels) ---
\node[src=cT, minimum width=0.45cm] (bdt) at (4.5,7.8) {\scriptsize$\Delta t$};
\node[src=cT, minimum width=0.8cm] (bfr) at (4.5,7.2) {\scriptsize Fourier};
\draw[cfl=cT, thin] (bdt) -- (bfr);

\node[src=cN, minimum width=0.45cm] (bz) at (4.5,6.45) {\scriptsize$z$};

\draw[cfl=cT] (bfr.east) -- (mt1.west);
\draw[cfl=cN] (bz.east) -- (mn1.west);

\draw[cT!70, line width=0.9pt, rounded corners=2pt]
    (5.2,7.2) -- (5.2,3.3) -- (mt2.west);
\draw[cN!70, line width=0.9pt, rounded corners=2pt]
    (5.0,6.45) -- (5.0,2.55) -- (mn2.west);
\fill[cT!70] (5.2,7.2) circle (0.05);
\fill[cN!70] (5.0,6.45) circle (0.05);

% --- Annotations ---
\node[font=\tiny\sffamily\bfseries, text=cT!85]
    at (10.4,7.2) {1\textsuperscript{st}};
\node[font=\tiny\sffamily\bfseries, text=cN!85]
    at (10.4,6.45) {2\textsuperscript{nd}};

\draw[decorate, decoration={brace, amplitude=2.5pt}, cG!65, line width=0.8pt]
    (10.7,7.35) -- (10.7,6.3);
\node[font=\tiny\sffamily, text=cG!80, anchor=west, align=left]
    at (10.8,6.82) {sequential\\[-1pt]modulation};

\node[ann, text=cM!65, font=\tiny\sffamily\itshape, anchor=east]
    at (6.55,7.85) {MSA};
\node[ann, text=cM!65, font=\tiny\sffamily\itshape, anchor=east]
    at (6.55,3.95) {MLP};

\node[ann, text=cX!55, anchor=west] at (10.25,2.6) {\tiny skip};

% ===== LEGEND =====
\draw[cX!45, line width=0.8pt] (-0.4,-0.35) -- (11.3,-0.35);

\fill[cT!14, draw=cT!65, rounded corners=1pt, line width=0.6pt]
    (0,-0.65) rectangle (0.4,-0.45);
\node[ann, anchor=west] at (0.5,-0.55) {Timestep conditioning ($\Delta t$)};

\fill[cN!12, draw=cN!65, rounded corners=1pt, line width=0.6pt]
    (4.2,-0.65) rectangle (4.6,-0.45);
\node[ann, anchor=west] at (4.7,-0.55) {Noise conditioning ($z$)};

\fill[cM!8, draw=cM!58, rounded corners=1pt, line width=0.6pt]
    (8.0,-0.65) rectangle (8.4,-0.45);
\node[ann, anchor=west] at (8.5,-0.55) {Standard layers};

\end{tikzpicture}

\caption{\textbf{Multi-timestep conditioning architecture.}
\textbf{(a)}~The model is a stack of $L$ transformer blocks, each receiving the
same conditioning signals.
\textbf{(b)}~Inside each block, the normalized activations are modulated
\emph{sequentially}: first by the Fourier-embedded timestep $\Delta t$
(amber, shift/scale/gate), then by ensemble noise $z$ (green,
shift/scale/gate).  This pattern applies identically to both the
neighborhood self-attention and MLP branches, with gates composing multiplicatively.
Highlighted bands mark every component that is conditioned---all
other layers (LayerNorm, Neighborhood Attention, MLP) are standard and unmodified.}
\label{fig:architecture}
\end{figure*}

\textbf{Implementing a modulation mechanism}. Implementation-wise, both $\Delta t$ and $z$ are routed through the same conditioning pathway via an adaptive layer norm (AdaLN-Zero). This is analogous to the combined timestep and class-label conditioning of Diffusion Transformers \citep{peebles2023scalable}. However, we modify the mechanism to use independent weights for each signal, and apply the modulations sequentially, in order to more explicitly separate their action (Fig.~\ref{fig:architecture}).

The scalar $\Delta t$ is first expanded into a fixed Fourier-feature representation and then linearly projected (per block) to produce the AdaLN parameters. The noise vector $z$ is processed analogously through its own projection. For a generic block activation $u$ (channel-wise), the AdaLN affine modulation takes the standard form
\(u \;\mapsto\; \bigl(1+\gamma\bigr)\odot \mathrm{LN}(u) + \beta,\) followed by a gated residual update of the form
\(u \;\mapsto\; u \;+\; g \odot F(\,\cdot\,),\) where $\mathrm{LN}(\cdot)$ is layer normalization, $F$ denotes the block sublayer (attention or MLP), and $(\beta,\gamma,g)$ are conditioning-dependent per-channel vectors. AdaLN-Zero initializes these modulations at (or near) zero, so the network begins close to an unmodulated residual block and learns to introduce conditioning smoothly.

We apply the timestep modulation first and the noise modulation second. Writing the timestep-derived parameters as $(\beta_{\Delta t},\gamma_{\Delta t},g_{\Delta t})$ and the noise-derived parameters as $(\beta_{z},\gamma_{z},g_{z})$, the affine part is composed sequentially:
\begin{equation}
\label{eq:sequential_adaln}
\tilde u
\;=\;
\bigl(1+\gamma_{z}\bigr)\odot\Bigl(\,\bigl(1+\gamma_{\Delta t}\bigr)\odot \mathrm{LN}(u) + \beta_{\Delta t}\Bigr) + \beta_{z}.
\end{equation}
This ordering makes $\Delta t$ the primary controller of the internal computation corresponding to the desired transition operator, while $z$ acts as a second-order perturbation around that operator. For the residual gate, we use multiplicative composition (implemented via a bounded parameterization). The sublayer output is then added to the residual stream. The two gates serve two purposes: $\Delta t$ provides a conditioning mechanism for adapting to the required transition scale, whereas $z$ works as an uncertainty injection mechanism for both optimizing the model and producing ensemble outputs given a chosen $\Delta t$. This sequential modulation is applied block-wise.

More broadly, AdaLN is a flexible mechanism that enables conditioning on any number of global signals that should coherently modulate the entire network rather than individual tokens. Other potential applications include climate scenarios or other external forcing terms, metadata of the initial conditions, or end-user forecast preferences such as calibration vs. sharpness.
\subsection{Mixed-Timestep Optimization}
\label{subsec:mixed_timestep_optimization}

Training seeks a single parameter set $\theta$ that yields accurate kernels across a range of timesteps, so that induced long-lead marginals remain accurate under rollout and under mixed-resolution rollouts. To avoid distribution shifts associated with staged curricula over timesteps, we train from scratch on a mixed-timestep objective: each training example samples $\Delta t$ uniformly from the supported set $\mathcal{T}$ and uses as target the next state $X_{t+\Delta t}$, taken $\Delta t$ hours after $X_t$.

A direct uniform mixture over the supported $\Delta t$ values creates an imbalance in gradient contributions because the typical scale of forecast uncertainty, and hence the typical magnitude of a proper scoring rule, increases with $\Delta t$. To address this imbalance, we use timestep-dependent loss scaling:
\begin{equation}
\label{eq:mixed_timestep_objective}
\mathcal{L}^*(\theta)
\;=\;
\mathbb{E}_{\Delta t \sim U(\mathcal{T})}\ 
\mathbb{E}\!\left[
\sqrt{\frac{\Delta t_{\mathrm{base}}}{\Delta t}}\,
\mathcal{L}\!\Big(q_\theta(\cdot \mid X_t,C,\Delta t),\, X_{t+\Delta t}\Big)
\right],
\qquad
\Delta t_{\mathrm{base}}=24\text{ h}.
\end{equation}

\textbf{Behavior at inference time}. At inference, the user supplies $\Delta t$ as an explicit input. This value is propagated through the AdaLN pathway at every step, so changing $\Delta t$ changes the invoked transition kernel $q_\theta(\cdot\mid X_t,C,\Delta t)$ without changing network weights or architecture, and the noise input $z$ continues to index ensemble members within that kernel.

\section{\method: A Global Model for Multi-Scale Forecasting}
\label{sec:global_model}
Having described the key multi-timestep inference mechanism, we now present the remaining details of \method, a global forecasting system that leverages this mechanism to provide a single model that can be deployed at varying temporal scales. \method is a successor to GEM-2 \citep{rauba2026probabilistic} and follows similar key design patterns: (i) it models both instantaneous prognostic variables and sub-timestep diagnostic accumulations natively; (ii) it uses a secondary spectral power CRPS loss; and (iii) it follows a similar validation pipeline. We briefly describe the remaining key architectural and training details below.

\subsection{Modeling in anomaly space}
\label{subsec:anomaly}

We parameterize a subset of the forecasting dynamics in anomaly space, i.e. a residual defined relative to a time-dependent climatology. Prior to training, we precompute a static climatology dataset from the training dataset, indexed by an hour-of-year coordinate. The climatology is lowpass filtered along the day-of-year coordinate to eliminate sampling noise from the finite historical record, giving a smoothly oscillating estimate of the seasonal and diurnal cycles.

The model architecture then shifts chosen variables into anomaly space by subtracting the climatology upstream of the core model backbone, and adding it back after. The optimization target then remains the full weather state, but the transformer backbone learns to evolve the residual state.

Specifically, \method shifts all temperature, geopotential, and pressure variables into anomaly space. We choose this subset of variables where the low-frequency seasonal cycle is closely associated with the external solar forcing. We intentionally do not transform variables for which the climatological cycle cannot be directly connected to solar heating, or for which hard physical bounds make accurate representation within the model's internal distributions challenging.

\subsection{Architecture and training}
\label{efficient_architecture}

\textbf{Backbone architecture}. \method uses a Neighborhood Attention Transformer (NAT) backbone operating on an equirectangular latitude-longitude grid. Inputs are patchified with a patch size of 4 pixels and linearly projected into a latent space, after which a stack of identical transformer blocks performs the core computation. Each block consists of (i) pre-norm of the input activations, (ii) 2D neighborhood self-attention with resolution-scaled local kernels and circular longitude padding to respect spherical topology, and (iii) a SwiGLU MLP. Attention employs QK normalization with a learned bounded per-head temperature applied to queries, and 2D rotary positional embeddings defined in resolution-normalized coordinates.  After the processor stack, a global RMS normalization stabilizes the accumulated residual stream. Decoding back to the pixel grid is performed through progressive upsampling, and a full-resolution skip connection concatenates the original input fields with decoded features before the output heads. Separate linear projection heads are used for prognostic and diagnostic outputs.

\textbf{Training data}. \method was trained exclusively on ERA5 data from 1979-2019. For its prognostic state, the model uses 5 variables on 13 pressure levels: temperature, geopotential, specific humidity, and u/v wind components, plus 2m temperature, skin temperature, and mean sea-level pressure. 13 static features are used only as model inputs to give additional per-token context, including different embeddings of space and time plus the ERA5 elevation and land-sea-mask grids. 23 additional timestep-accumulations, primarily of surface properties, are emitted as diagnostic outputs.

\textbf{Optimization}. We optimize most of the transformer backbone parameters with the Muon optimizer \citep{liu2025muon, jordan2024muon}. A smaller subset, empirically found to be unstable under Muon, is instead evolved with AdamW updates. The model is trained in multiple stages: a single-step base stage, followed by any number of rollout stages. In the rollout stages, model outputs are autoregressively fed back as input for the next step, and the loss is evaluated over the full length of the rollout. Rollout training is performed with the same mixed-timestep procedure as the base training, although timestep samples can be preferentially weighted towards shorter timesteps since these demonstrate the largest stability gains from rollout training.

\textbf{Loss function.} The training objective combines two complementary scores: a per-gridpoint fair CRPS and a spectral CRPS computed in the spherical harmonic domain. Given $S$ ensemble samples $\{\hat{x}^{(s)}\}_{s=1}^{S}$ and target $x$, the fair CRPS is $\mathcal{L}_{\mathrm{CRPS}} = S^{-1}\sum_s |\hat{x}^{(s)} - x| - [2S(S-1)]^{-1}\sum_{s,s'}|\hat{x}^{(s)} - \hat{x}^{(s')}|$, evaluated pointwise and weighted by $w(\phi) = \max(\cos\phi\,/\,\overline{\cos\phi},\; 0.1)$ to account for spherical geometry. For the spectral term, we compute the degree-averaged power spectrum $P_\ell(x)$ via the real spherical harmonic transform, apply a power-law emphasis $(\ell+1)^\gamma$ ($\gamma\!=\!1$) with a sigmoid taper above $\ell_{\mathrm{tap}} = \lfloor L_{\max}/2 \rfloor$, and evaluate the fair CRPS on the log-compressed spectra: $\mathcal{L}_{\mathrm{SH}} = \mathcal{L}_{\mathrm{CRPS}}\!\big(\log(1+\tilde{P}_\ell(\hat{x})),\;\log(1+\tilde{P}_\ell(x))\big)$. Both losses are computed independently for prognostic and diagnostic output heads, similar to GEM-2.

\begin{table}[H]
\vspace{0.8em}
\newcommand{\ymark}{\textcolor{black!80}{\cmark}}
\newcommand{\nmark}{\textcolor{black!25}{\xmark}}
\centering
\scriptsize
\renewcommand{\arraystretch}{1.48}
\setlength{\tabcolsep}{4pt}
\setlength{\abovecaptionskip}{0.7em}
\begin{tabularx}{\textwidth}{@{}>{\hsize=.78\hsize\raggedright\arraybackslash}X c c c c >{\hsize=1.22\hsize\raggedright\arraybackslash}X c c@{}}
\toprule
\textbf{Model} &
\makecell{\textbf{Surface}\\\textbf{Fields}} &
\makecell{\textbf{Inference}\\$\boldsymbol{\Delta t}$} &
\textbf{Architecture} &
\textbf{Target} &
\textbf{Loss} &
\makecell{\textbf{Hidden}\\\textbf{dim.}} &
\makecell{\textbf{\#}\\\textbf{Params}} \\
\midrule

GenCast \citep{price2023gencast} &
6 & 12 & Graph & Time Residual & Diffusion & 512 & $\sim$57M\\

FGN \citep{alet2025skillful} &
6 & 6 & Graph & State & CRPS & 768 & $\sim$720M\\

AIFS-ENS \citep{lang2024aifs} &
$\sim$17 & 6 & Graph & Time Residual & CRPS & 1024 & $\sim$229M\\

FourCastNet-3 \citep{bonev2025fourcastnet3} &
6 & 6 & Spherical Harmonics & State & CRPS \& SHT Components & 641 & $\sim$710M\\

Atlas \citep{kossaifi2026demystifying} &
9 & 6 & Rectangular & Latent Residual & CRPS \& SHT Components & 3328 & $\sim$4.3B\\

\addlinespace[3pt]
\rowcolor{blue!5}
\textbf{\method\ (ours)} &
23 & Configurable & Rectangular & Anomaly/State & CRPS \& Spectral Power & 768 & 134M\\

\bottomrule
\end{tabularx}
\caption{\textbf{Comparative design-space summary for selected probabilistic ML weather models.}
Surface Fields reports the number of native surface-level variables emitted by the model beyond the 3-dimensional prognostic state. Inference~$\Delta
t$ indicates the autoregressive timestep. Architecture identifies the geometry of the model backbone.
Target describes the primary forecast representation learned by the
model, while Loss lists the main probabilistic training objective. Hidden dim.\
denotes the primary backbone width.}
\label{tab:probabilistic_comparison}
\end{table}

\section{Experimental results}
\label{sec:results}

\method is evaluated in two parts. We first consider a single operational 0.25$^\circ$ resolution model trained on a mixed 6\,h and 24\,h timestep distribution, where the goal is to establish the overall performance envelope of the architecture against NWP and other ML baselines. We then study a set of controlled ablations at 1$^\circ$ resolution to further explore the performance characteristics of different variants of the architecture.

\textbf{Operational model.} The operational \method model was trained in two stages. All training was done at an effective batch size of 128, using gradient accumulation to scale consistently across different spatial resolution experiments and rollout lengths. The base training on single-step forecasts was run for a total of $\sim$70{,}000 optimizer steps using an equal mix of 6\,h and 24\,h timesteps. In the second phase, the model was fine-tuned on a rollout length of 4 for $\sim$7{,}000 optimizer steps. Timestep samples were preferentially weighted towards 6\,h forecasts during rollout training. With activation checkpointing the model fits on a single GPU, hence only data parallelism was applied to train the model across 16 H100s. Total training cost was $\sim$250 H100-days, roughly half the cost of a single FGN seed and one-quarter the cost of AIFS-ENS.

All forecasts and reforecasts after July 2024 are initialized with the ECMWF IFS analysis state, and reforecasts prior to that with ERA5. In direct comparisons, we found initializing with ERA5, which uses a broader assimilation window, leads to a moderate skill enhancement for surface variables in the first $\sim$2--3 days of the forecast, but quickly decays beyond that, and has almost no impact on upper-level variable scores such as z500. All validation is performed using ERA5 as the source of truth, and global scores are averaged using cosine-latitude weights on the equirectangular grid.

\begin{figure*}[!htb]
\centering
\includegraphics[width=0.98\textwidth]{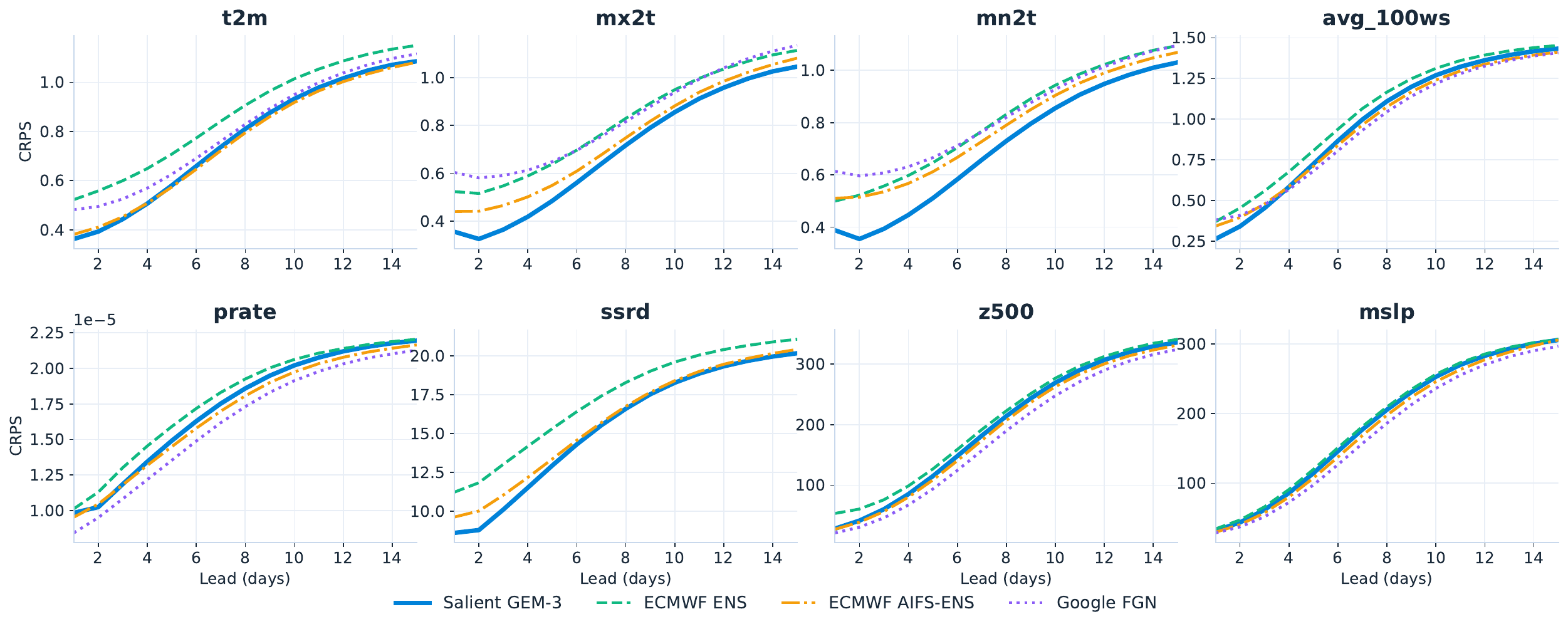}
\captionof{figure}{\textbf{Operational model medium-range CRPS.} Global CRPS over the first 15 forecast days for the operational \method model against ECMWF ENS, ECMWF AIFS-ENS, and Google FGN. Evaluated on 242 initializations from July 2025 to March 2026, the period for which AIFS-ENS forecasts are available.}
\label{fig:production_mr}

\vspace{0.5em}

\includegraphics[width=0.98\textwidth]{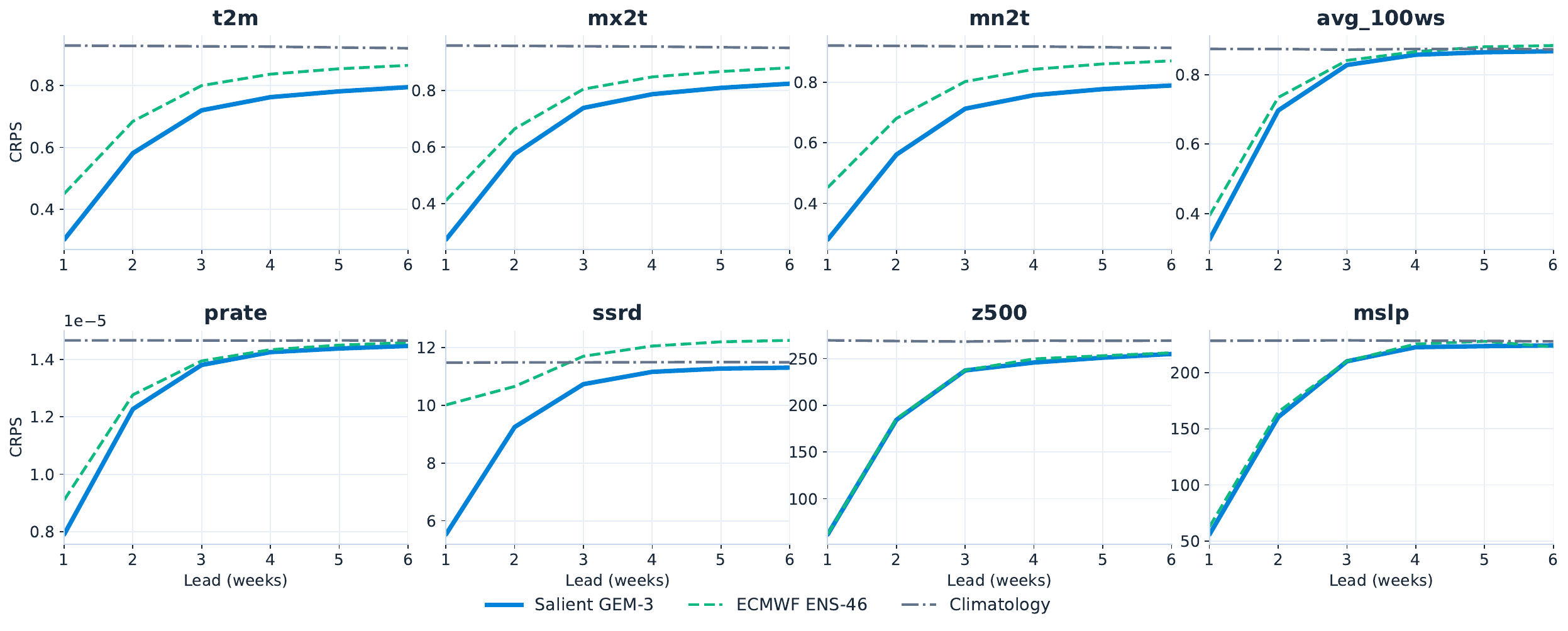}
\captionof{figure}{\textbf{Operational model extended-range CRPS.} Global CRPS through 6 weeks for the operational \method model, ECMWF ENS-46, and a 30-year climatological baseline. Evaluated on 104 initializations from July 2023 to July 2025.}
\label{fig:production_s2s}
\end{figure*}

On standard 15-day global CRPS benchmarks, \method consistently outperforms ECMWF ENS and is competitive with the latest ML systems (Figure~\ref{fig:production_mr}). \method is roughly on par with AIFS-ENS across most variables, while FGN outperforms on 500 hPa geopotential, mean sea-level pressure, and precipitation, but underperforms on 2m temperature. Because the other ML models do not natively output minimum and maximum temperature, we have simply estimated these from 6-hourly snapshots here. \method therefore has a structural advantage on these variables, leading to the larger apparent skill gains.

\method also displays stable and skillful extended-range forecasts (Figure~\ref{fig:production_s2s}), beating or equaling the ECMWF-ENS sub-seasonal forecasts out to 46 days, and maintaining positive or neutral skill over climatology across this horizon. \method is also generally well-behaved under longer rollouts, and the operational forecasts are run to 126 days. While the 6\,h timestep inference has generally good stability beyond 15 days, the coarser 24\,h timestep demonstrates better scores on extended rollouts due to the reduced accumulation of error and subsequent drift. This is the key practical benefit of the multi-timestep inference mechanism: a single model can utilize an optimal timestep across a wide range of forecast horizons.

\FloatBarrier
\textbf{Ablations on multi-timestep inference.} We additionally study a set of ablations at 1$^\circ$ resolution in order to further explore the parameter space. In this section we refer to models trained on multiple timesteps as ``generalists'', and models trained on a single timestep as ``specialists''. These models use an identical architecture and training recipe to the operational model, with the exception of the coarser spatial resolution, and a slightly modified training recipe weighted more heavily towards rollout training. The base single-step stage was run for 50{,}000 optimizer steps, followed by an additional 10{,}000 steps at rollout length 4, and a final 2{,}000 steps at rollout length 12. This modification was made in an attempt to improve the rollout stability of the sub-6\,h timesteps, which are particularly prone to drift under rollout. All ablation experiments were evaluated for 37 initialization dates at 10-day spacing for the year 2025.

We first probe the range of useful timesteps by extending the 6/24\,h mixed-timestep model to a broader generalist trained on 1/3/6/12/24\,h timesteps. This model is evaluated on forecasts run across all training timesteps to compare performance characteristics. Forecasts from 3--24\,h timesteps show broadly similar performance across the 15-day rollout, with slightly higher predictive power of the 3 and 6\,h timesteps relative to the 12 and 24\,h timesteps in the first week. The 1\,h forecast displays significant drift and error accumulation beginning already at day 2.

\begin{figure*}[!htbp]
\centering
\includegraphics[width=0.98\textwidth]{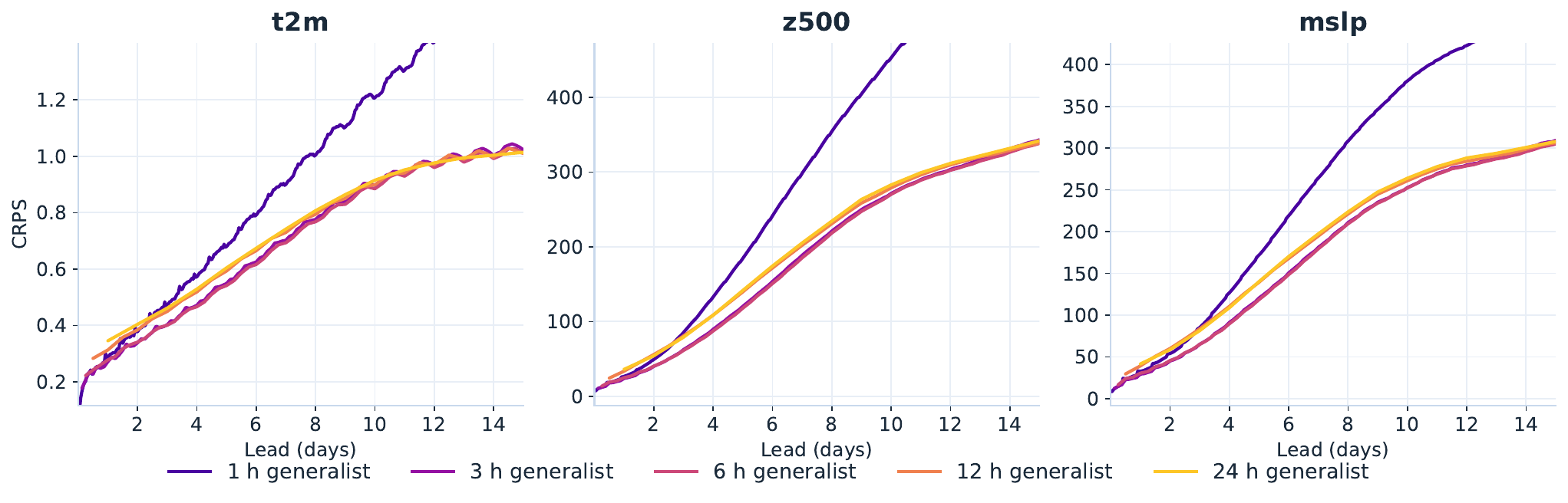}
\caption{\textbf{Performance of 1--24\,h timestep-generalist model.} CRPS evaluation of a single timestep-conditioned model evaluated at 1, 3, 6, 12, and 24\,h forecast timesteps.}
\label{fig:generalist_timestep_sweep}
\end{figure*}

\begin{figure*}[!htbp]
\centering
\includegraphics[width=0.72\textwidth]{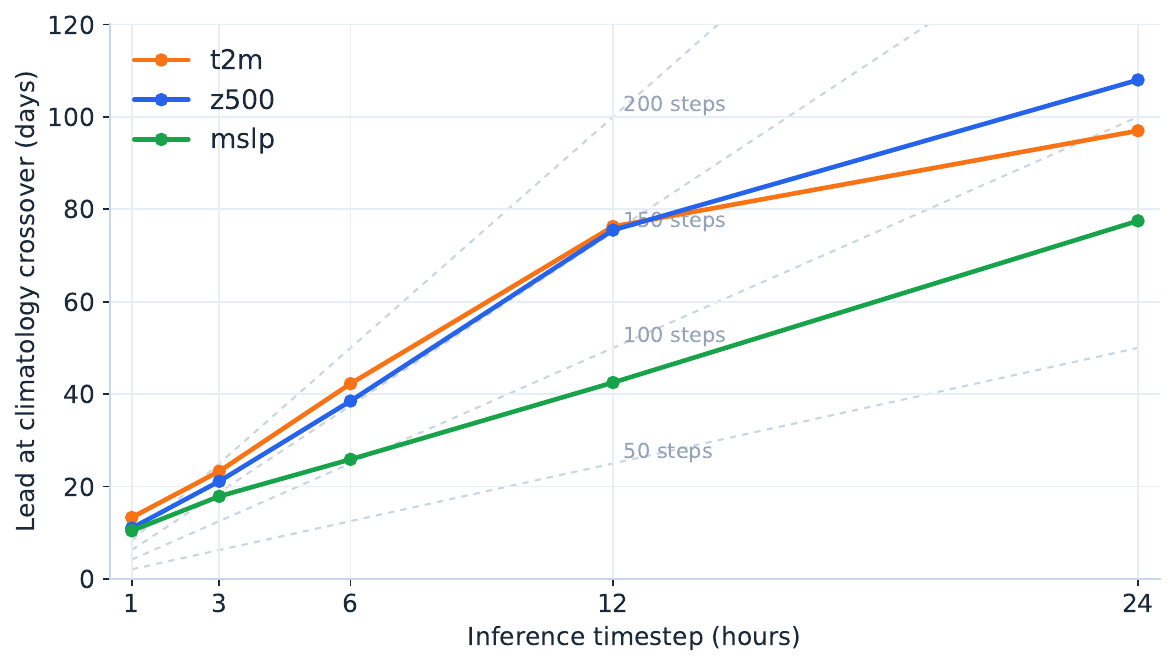}
\caption{\textbf{Degradation of the 1--24\,h generalist by timestep.} The lead time in days at which forecast CRPS first becomes worse than a climatological baseline, for each inference timestep of the generalist model. Faint guide lines indicate the equivalent number of autoregressive steps required to reach that crossover. Coarser timesteps remain skillful to later physical leads, while finer timesteps tolerate more recurrent applications before degrading to worse than climatology.}
\label{fig:generalist_crossover}
\end{figure*}

A useful stability metric is the lead time at which a forecast becomes worse than a climatological baseline. In Figure~\ref{fig:generalist_crossover} we compare the generalist model to a climatological baseline and compute the crossover lead time where globally averaged CRPS becomes worse than the baseline. Unsurprisingly, the crossover lead time increases with the inference timestep, implying that shallower rollout depths mitigate autoregressive error accumulation. In this configuration, the 24\,h forecast maintains positive skill out to about 3 months, while the 1\,h forecast degrades to worse than climatology by only 10 days. We can also compute the number of autoregressive steps required to reach the crossover lead time. In this framework, it is evident that the shorter timestep models can in fact be rolled out for a larger number of steps before degrading to worse than climatology. Taken together, this suggests that finer timesteps inject less error per update because they approximate an easier local transition, but not nearly enough to offset the much larger rollout depth needed to reach the same forecast lead.

\begin{figure*}[!htbp]
\centering
\includegraphics[width=0.98\textwidth]{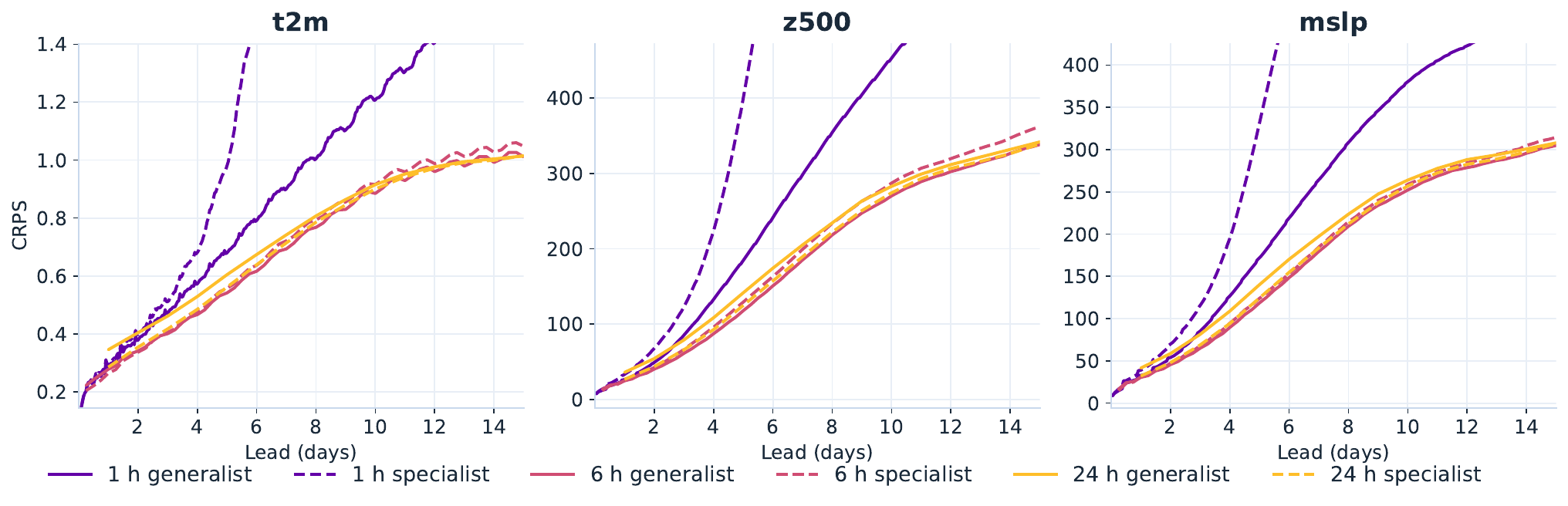}
\caption{\textbf{Performance of timestep-generalist versus timestep-specialists.} CRPS evaluation of forecasts run from the 1--24\,h timestep-generalist model against matched timestep-specialists trained at 1, 6, and 24\,h.}
\label{fig:generalist_vs_specialist}
\end{figure*}

We then compare the timestep-generalist model against a series of timestep-specialist models trained at 1, 6, and 24\,h (Figure~\ref{fig:generalist_vs_specialist}). The main result is that the generalist model run at a fine 1\,h timestep, while still prone to drift in a multi-day rollout, has much lower error accumulation compared to the specialist. This suggests a regularization effect of multi-timestep training which improves stability of short-step forecasts. The 6\,h generalist also shows very slight improvements in stability relative to the specialist. On 1 and 6\,h steps, the generalist shows very slightly worse performance in the first few steps, but in a forecast rollout this is offset by the lower error accumulation; the 6\,h generalist and specialist demonstrate nearly identical performance across the 15-day horizon.

In contrast, the 24\,h specialist actually shows better performance on the first $\sim$7 days, and then converges to the same performance as the generalist. This suggests that longer timesteps benefit from a more specialized transition operator, and that there is a limit to the utility of weight-sharing above a certain timestep length. We note that the 24\,h specialist nearly matches the performance of the higher-resolution 6\,h forecasts, meaning a relatively coarse forecast timestep can still produce skillful and competitive forecasts in the medium-range.

\begin{figure*}[!htbp]
\centering
\includegraphics[width=0.98\textwidth]{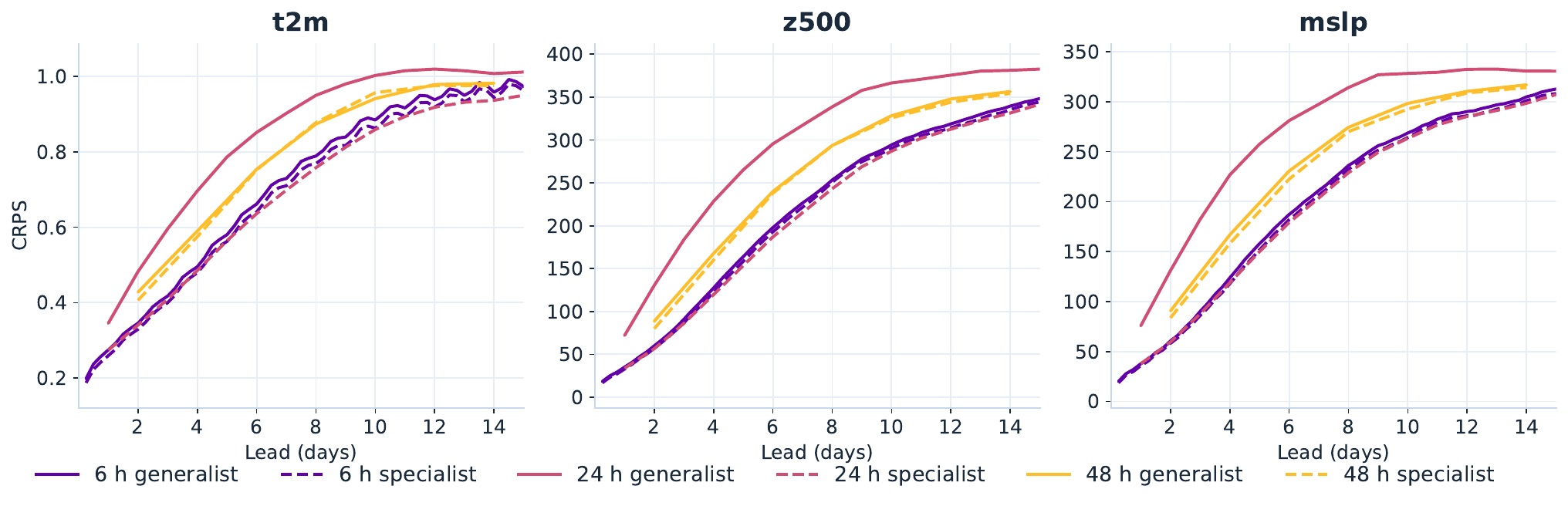}
\caption{\textbf{Performance of a 6--48\,h timestep-generalist model.} CRPS evaluation of forecasts run from a 6--48\,h timestep-generalist model against matched 6, 24, and 48\,h timestep-specialists.}
\label{fig:timestep_limit}
\end{figure*}

Next, we attempt to push the upper limit of the timestep range by training a generalist on 6/24/48\,h timesteps, and also comparing to equivalent specialists (Figure~\ref{fig:timestep_limit}). Here we find that all 48\,h step models perform significantly worse than 24\,h steps, and also that the inclusion of a 48\,h step degrades the generalist model overall, with the 24\,h step generalist in particular showing poor performance relative to other models. This suggests that a 48\,h step spans too much dynamical evolution of the system, and a single learned transition operator cannot adequately represent the 48\,h step in the present architecture. Combined with the slight degradation of the 24\,h generalist relative to the 24\,h specialist, this result also suggests that limitations of our timestep conditioning scheme begin to manifest around the 24\,h mark, where a low-dimensional modulation of shared weights is no longer sufficient to optimally forecast across such a wide range of step sizes.

\begin{figure*}[!htbp]
\centering
\includegraphics[width=0.98\textwidth]{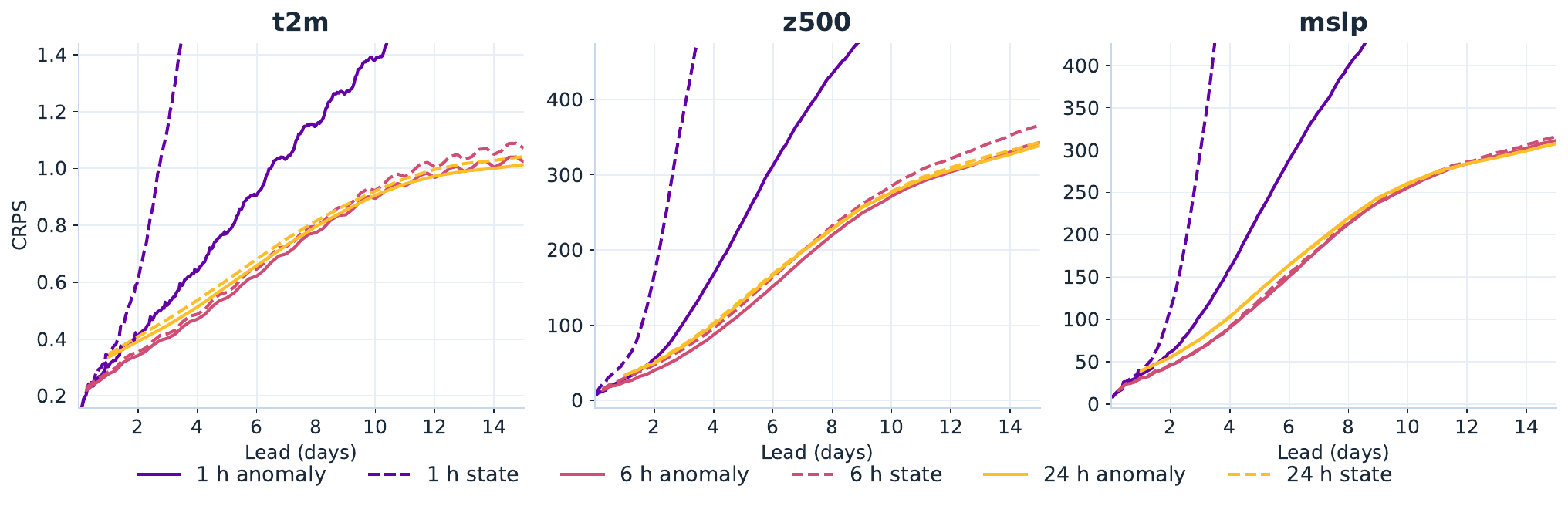}
\caption{\textbf{Anomaly versus state forecasting.} Comparison on three core variables across the useful 1--24\,h timestep range. Colors encode timestep, with solid lines for anomaly forecasting and dashed lines for state forecasting. Selective anomaly modeling most clearly improves the short-step regime while leaving the broader 6--24\,h behavior largely unchanged.}
\label{fig:anomaly_vs_state}
\end{figure*}

\textbf{Anomaly-space modeling ablation.} Finally, we perform a comparison of the selective anomaly-space modeling formulation (Figure~\ref{fig:anomaly_vs_state}). Here we compare generalist models trained on the 1/3/6/12/24\,h timestep set, one using the mixed anomaly-space formulation and one using a standard state-space formulation. The results demonstrate that the anomaly-space formulation is beneficial for rollout stability across all timesteps, but particularly so for the shorter 1 and 6\,h steps where autoregressive error accumulation is accelerated by the rollout depth. The initial step scores are effectively identical between both formulations.

\begin{figure*}[!htbp]
\centering
\includegraphics[width=0.98\textwidth]{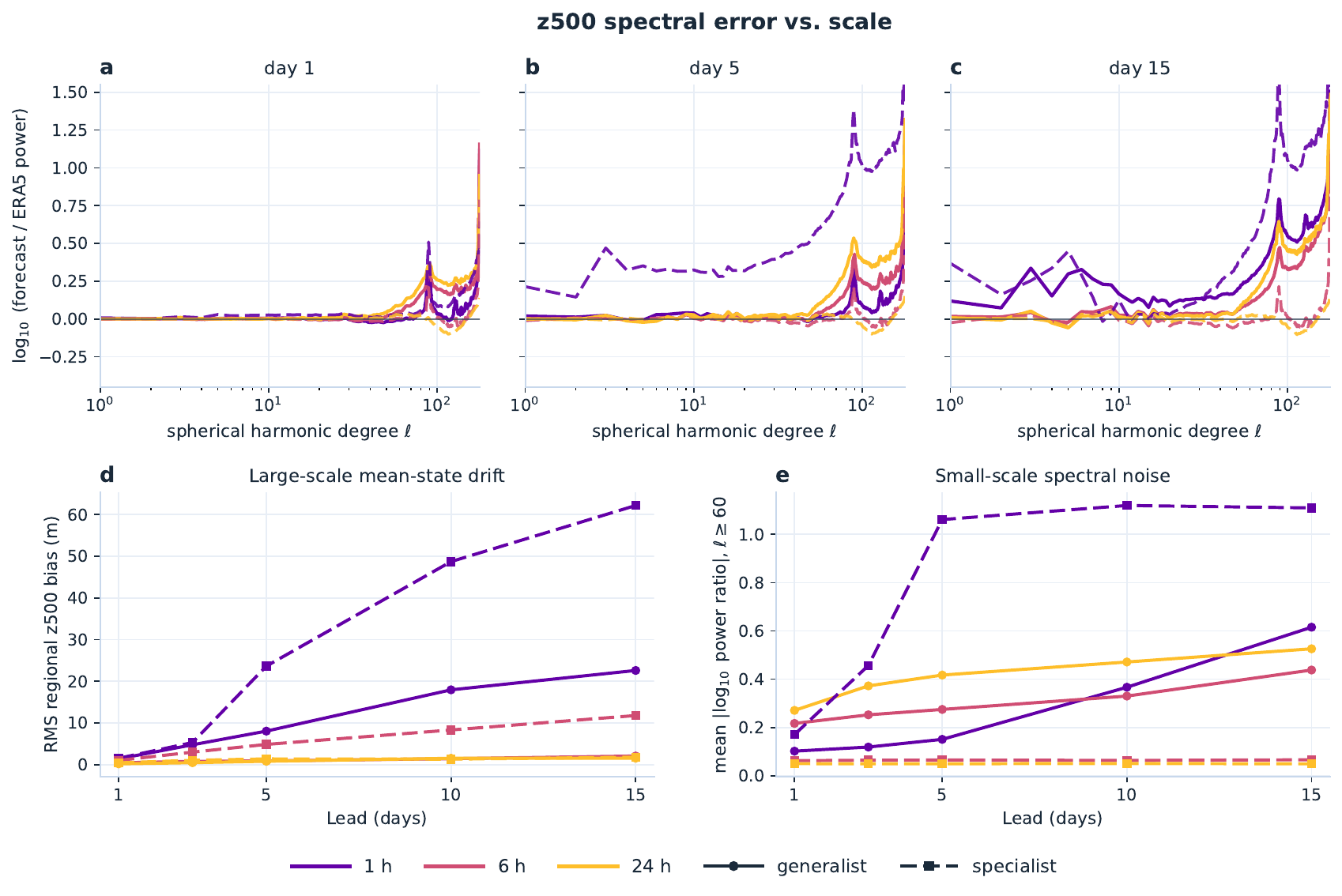}
\caption{\textbf{Mechanisms of rollout degradation.} Additional error diagnostics for the 1--24\,h timestep-generalist and matched 1, 6, and 24\,h specialists. \textbf{(a--c)}~$\log_{10}$ ratio of forecast to ERA5 degree-averaged z500 spherical harmonic power spectra at 1, 5, and 15-day lead; values above zero indicate excess power. \textbf{(d)}~Large-scale mean-state drift: RMS of the regional-mean z500 bias averaged over three latitude bands. \textbf{(e)}~Small-scale spectral distortion: degree-averaged absolute log-power ratio for $\ell \geq 60$.}
\label{fig:spectral_degradation}
\end{figure*}

\textbf{Spectral fidelity and mean-state drift under rollout.} Aggregate CRPS summarizes skill but obscures \textit{how} a rollout degrades. To probe this, we decompose z500 rollout degradation into two categories: a large-scale \textit{mean-state drift}, quantified as the root-mean-square of the regional-mean z500 bias over three latitude bands (NH extratropics, tropics, and SH extratropics); and a small-scale \textit{spectral distortion}, quantified as the degree-averaged absolute $\log_{10}$ ratio of forecast to ERA5 power at high wavenumber ($\ell \geq 60$). Figure~\ref{fig:spectral_degradation} tracks both channels for the 1--24\,h generalist and matched 1, 6, and 24\,h specialists. At 6 and 24\,h steps, the specialists track the ERA5 spectrum very closely and accumulate negligible small-scale distortion, whereas the generalist injects some excess small-scale power that compounds with lead. The ordering reverses for mean-state drift: the generalist effectively suppresses drift, particularly at 1 and 6\,h steps.

These diagnostics expose an interesting trade-off in the timestep-conditioning mechanism. A single set of weights, modulated only by $\Delta t$, must serve steps that require very different amounts of small-scale dissipation per update---almost none at 1\,h, substantially more at 24\,h. The shared operator appears to settle on a compromise: enough effective dissipation to arrest the drift and grid-scale noise that would otherwise destabilize fine-step rollouts, but too little for coarse steps, leaving a residual excess of small-scale variance that accumulates over the rollout.

Mixed-timestep training therefore trades a modest, slowly accumulating loss of small-scale spectral fidelity for a large reduction in mean-state drift. This cost is only weakly reflected in aggregate CRPS, and is largest for the wide 1--24\,h mixture shown here. The 6/24\,h operational model spans a narrower timestep range and likely shows less spectral distortion relative to equivalent specialist models. We also note that small-scale spectral quality is highly dependent on architecture details. The patchification approach used here is inherently susceptible to patch-scale artifacts relative to more sophisticated encoder-decoder designs. Larger embedding dimensions can also improve small-scale spectral fidelity, as capacity-limited models preferentially learn large-scale dynamics due to their larger contribution to standard loss functions.

\section{Discussion}
\label{sec:discussion}

The ablations above raise several interesting questions about the behavior of this architecture, which we address here.

\textbf{Why low-dimensional timestep conditioning works.} Within a broad timestep range, empirically determined to be $\leq 24$\,h, AdaLN-style conditional modulation effectively enables weight-sharing across the primary transformer stack, with only low-dimensional modulation required to adjust the transition operator for each timestep. Within this regime, the learned transition appears regular enough that a single set of latent operations can be attenuated or amplified as needed to represent an appropriate transition for each timestep. We can think of changing $\Delta t$ as asking the model to advect structures farther or apply growth or dissipation processes appropriate to the timestep interval.

\textbf{Why generalists stabilize rollouts.} The generalist-specialist comparison on short-step inference (Figure~\ref{fig:generalist_vs_specialist}) shows a consistent pattern: at 1 and 6\,h, the generalist accumulates error more slowly over multi-day rollouts, even though it is slightly worse on the first few steps. Mixed-timestep training evidently acts as a weak regularizer, producing a transition operator less prone to destabilizing error accumulation. The error decomposition shown in Figure~\ref{fig:spectral_degradation} suggests that this stabilizing effect is primarily due to a reduction in mean-state drift, with more mixed results for spectral realism. The generalist model has better and more stable short-step spectra, but worse coarse-step spectra. A single $\Delta t$-conditioned operator evidently cannot reproduce the scale-dependent response of every specialist at once---it damps the large-scale imbalances that destabilize fine-step rollouts, while under-representing the small-scale dissipation that keeps the coarse-step specialists spectrally faithful \citep{lehmann2026longrollouts}. In all experiments shown here, rollout training also improves short-step stability, yet it does not fully remove depth-dependent error growth and becomes costly at the rollout lengths needed for very fine timesteps.

\textbf{Why the useful timestep regime is broad but not unbounded.} The ablations suggest that there is no single optimal timestep for medium-range forecasting within a useful range of 3--24\,h, but they do show failure modes at either end of the spectrum. Within the optimal range, two effects plausibly offset one another: finer timesteps resolve short-range evolution more faithfully, while coarser timesteps reduce the number of autoregressive iterations needed to reach a given forecast horizon. Below this range, e.g. 1\,h timesteps, errors accumulate rapidly in units of physical time due to the large rollout depth. Above it, e.g. 48\,h timesteps, the one-step operator must memorize highly non-linear evolution and large state jumps, including multiple realizations of the diurnal cycle. It is possible that a larger parameter-count model, or one with global attention, could extend the range of skillful timesteps, but this would cancel out the computational efficiency gains of coarser timestep inference.

\textbf{Timestep as an operational choice.} This broad plateau suggests that the choice of timestep in ML forecasting systems is not only one of skill, but rather a more subtle balance of usability, cost, and rollout stability. Finer steps are inherently valuable when sub-daily temporal detail is desired, but they come at the cost of deeper rollouts, more compute, and greater storage. Coarser steps are attractive in the extended range because they reduce rollout depth with similar synoptic skill. \method's architecture, which includes flexible and native sub-timestep diagnostic outputs, further relaxes this trade-off, because even a coarse timestep model can output key sub-timestep statistics like minimum and maximum temperature or wind gusts. Taken together, these considerations motivate the configurable inference scheme chosen for the operational \method model, which resolves diurnal processes in the high-predictability 14-day forecast horizon with a 6\,h timestep, then switches to a 24\,h timestep for stability in the extended range rollout.

\textbf{Dynamics in anomaly-space.} The anomaly-space formulation is empirically shown here to be beneficial to average forecast skill and stability. It also raises questions about the extent to which the model is still approximating physical equations, since part of the transition operator is now working with non-physical, transformed quantities. With this in mind, we choose to limit the anomaly-space transformation to specific quantities where, to first order, the transition between any two states can be approximately separated into one driven by synoptic dynamics and one driven by the external solar forcing. We postulate that this is approximately true for temperature, pressure, and geopotential, where the solar forcing imposes a quasi-static low-frequency structure that can be isolated from the active dynamics. This is not the case for other atmospheric properties like moisture and wind, whose seasonal and diurnal cycles are also connected to the external solar forcing, but through intermediate, non-linear processes. The anomaly-space formulation is also a poor fit for fields with hard lower or upper bounds (e.g., precipitation), because a time-dependent climatology forces continuous shifts in these saturation values.

\section{Conclusion}
\label{sec:conclusion}

\method introduces explicit timestep conditioning into a probabilistic global weather model, so that the rollout schedule becomes a configurable inference-time choice rather than a fixed property of the architecture. This makes it possible to use one parameter set at fine timesteps where short-range detail is most valuable, and at coarser timesteps where extended-range rollout depth, cost, and stability become the dominant concerns. This yields a model that outperforms ECMWF ENS on both medium- and extended-range benchmarks, remains competitive with current leading ML models on probabilistic medium-range benchmarks, and also emits native sub-timestep diagnostic quantities across the flexible timestep forecasts.

Controlled ablations show that the benefit is not only flexible inference scheduling. Mixed-timestep training also improves the rollout stability of short-step forecasts relative to matched timestep specialists, and combines well with selective anomaly-space modeling to reduce drift. This stability is not entirely free, however: the shared timestep-conditioned operator injects a small excess of high-wavenumber power that accumulates under rollout, exchanging a modest loss of small-scale spectral fidelity for a substantial reduction in large-scale mean-state drift. These experiments also identify the approximate bounds of the useful timestep range. Very small timesteps become increasingly vulnerable to drift under practical rollout lengths. Very large timesteps suffer from degraded predictive skill, and are not well-behaved under the proposed timestep-modulation scheme.

Taken together, these results support multi-timestep inference as a practical design for probabilistic weather forecasting. Within an appropriate timestep range, one model can cover both short-range and extended-range deployment settings by changing its inference schedule rather than its architecture.

\textbf{Open data}. A large public reforecast dataset from the \method model has been made publicly available. The dataset covers daily initializations from 2020--2025, with additional weekly and monthly extended-range forecasts from 2000--2019. It covers 16 variables with 50 ensemble members at each initialization. Access instructions and additional details are available at \url{https://gemv3.salient-open-data.com}.

\clearpage
\bibliographystyle{abbrvnat} % or plainnat/unsrtnat
\bibliography{library}        % matches your library.bib

% \appendix

% \newpage

% \section{Additional ablations}

% \begin{center}
% \includegraphics[width=0.98\textwidth]{images/quick_24h_architecture_ablation.pdf}
% \captionof{figure}{\textbf{24\,h specialist optimizer and backbone ablation.} Comparison of three 24\,h specialist models: the architecture described in the main text using NATTEN attention blocks and a hybrid-Muon optimizer, an AdamW variant with the same NATTEN backbone, and a Swin variant with the same hybrid-Muon optimizer and embedding dimension (768) and depth (12). Models were trained for a minimal 5,000 optimizer steps. Muon optimization leads to nearly 2x faster convergence than AdamW for a given forecast performance level, and the NATTEN backbone significantly outperforms the Swin variant, although further testing showed that this performance gap can be partially recovered with a larger embedding dimension and additional training time.}
% \label{fig:quick_24h_architecture_ablation}
% \end{center}

\end{document}